\documentclass[10pt,twocolumn,letterpaper]{article}
\usepackage{authblk}

\usepackage{cvpr}              

\usepackage{graphicx}
\usepackage{amsmath}
\usepackage{amssymb}
\usepackage{booktabs}
\usepackage{algorithm}
\usepackage{bm}
\usepackage{algpseudocode}
\usepackage{multirow}
\usepackage{enumitem}
\makeatletter
\@namedef{ver@everyshi.sty}{}
\makeatother
\usepackage{tikz}

\usepackage[pagebackref,breaklinks,colorlinks]{hyperref}

\usepackage[capitalize]{cleveref}
\crefname{section}{Sec.}{Secs.}
\Crefname{section}{Section}{Sections}
\Crefname{table}{Table}{Tables}
\crefname{table}{Tab.}{Tabs.}

\def\cvprPaperID{8620} 
\def\confName{CVPR}
\def\confYear{2023}

\newcommand{\todocite}[1]{\textcolor{blue}{Citation needed []}}

\newcommand{\ignorethis}[1]{}

\makeatletter
\DeclareRobustCommand\onedot{\futurelet\@let@token\@onedot}
\def\@onedot{\ifx\@let@token.\else.\null\fi\xspace}

\def\eg{{e.g}\onedot}

\makeatother

\newcommand{\bbR}{{\mathbb{R}}}

\newcommand{\bx}{\mathbf{x}}

\newcommand{\bv}{\mathbf{v}}

\newcommand{\btheta}{\bm{\theta}}

\newcommand{\bphi}{\boldsymbol{\phi}}

\begin{document}
\twocolumn[{%
\renewcommand\twocolumn[1][]{#1}%
\title{QFF: Quantized Fourier Features for Neural Field Representations}
\author[1]{Jae Yong Lee}
\author[1]{Yuqun Wu}
\author[2]{Chuhang Zou\footnote{The work does not related to the author’s position at Amazon.}}
\author[1]{Shenlong Wang}
\author[1]{Derek Hoiem}
\affil[1]{University of Illinois at Urbana-Champaign}
\affil[2]{Amazon.com}

\maketitle
\begin{center}

\vspace{-1em}
    \centering
    \scriptsize
    \captionsetup{type=figure}
    \begin{tabular}{@{}c@{}c@{}c@{}c@{}}
        \includegraphics[trim={4cm 4cm 4cm 4cm},clip,width=0.24\textwidth]{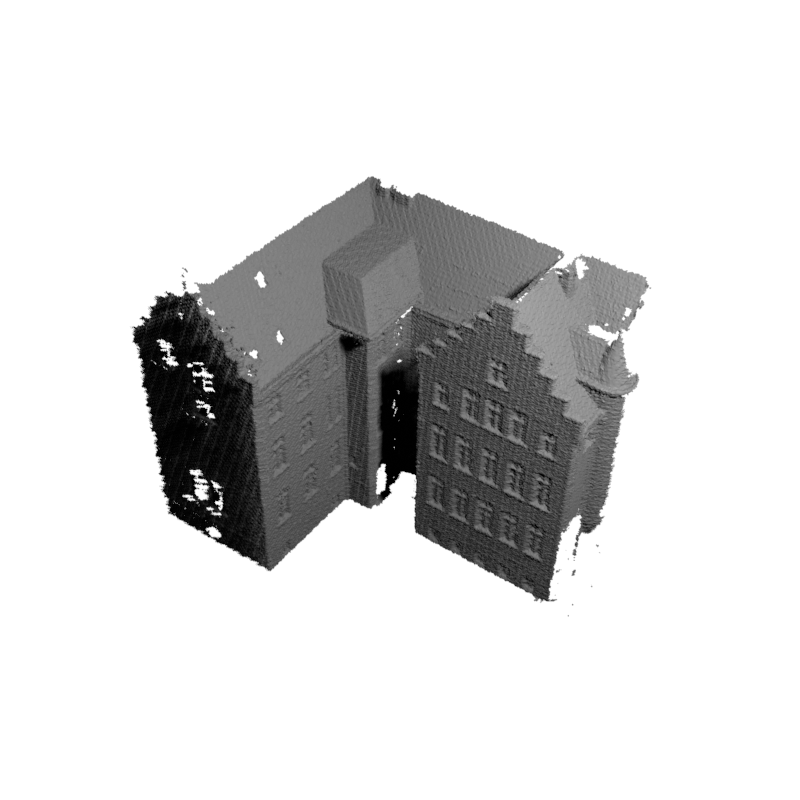} &
        \includegraphics[trim={4cm 4cm 4cm 4cm},clip,width=0.24\textwidth]{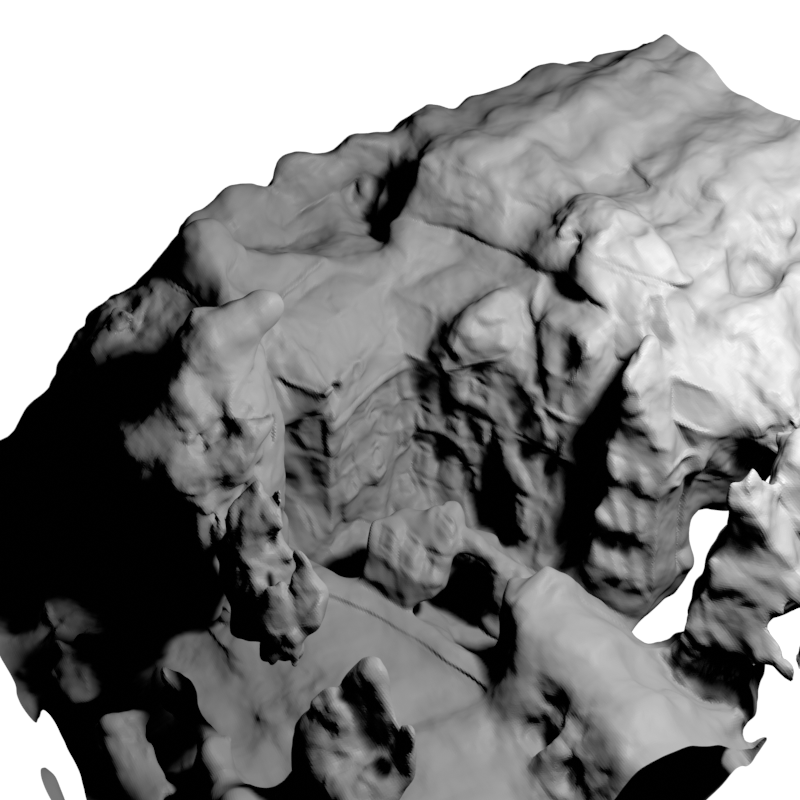} &
        \includegraphics[trim={4cm 4cm 4cm 4cm},clip,width=0.24\textwidth]{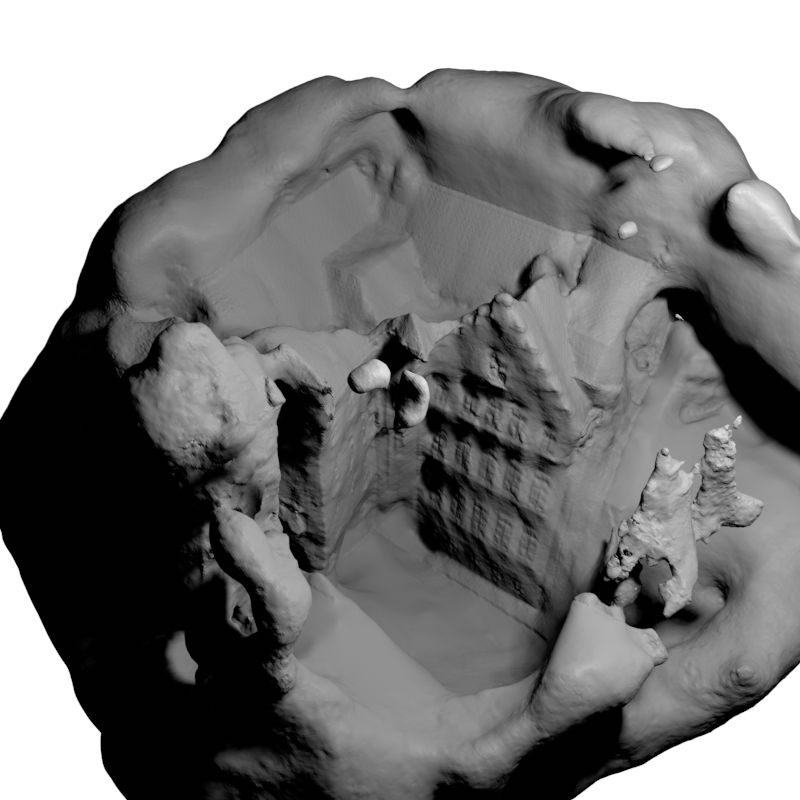} &
        \includegraphics[trim={4cm 4cm 4cm 4cm},clip,width=0.24\textwidth]{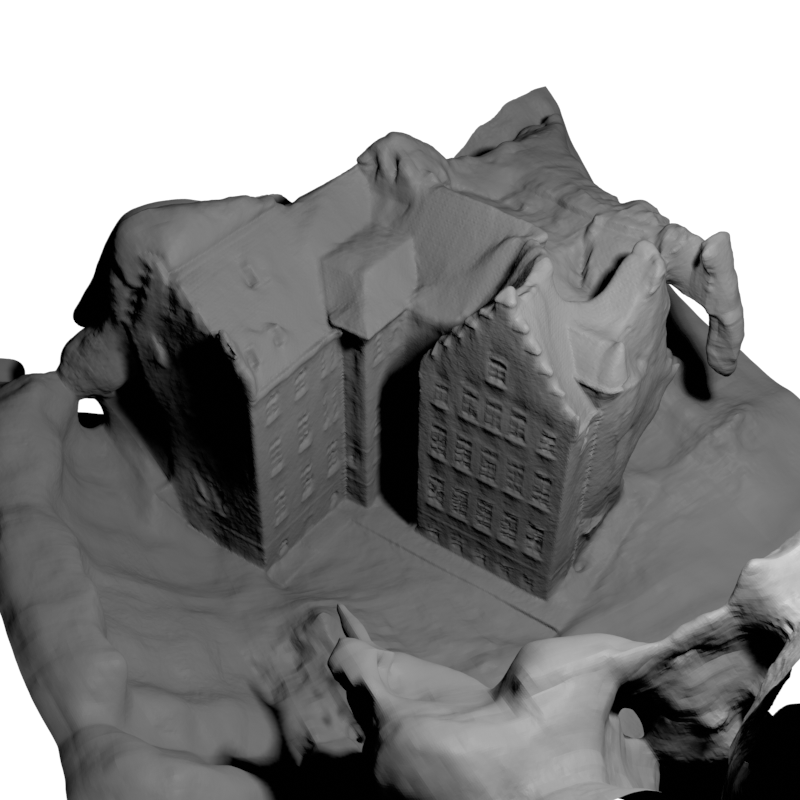}\\
        Ground Truth & MLP~\cite{yu2022monosdf} & Multi-Res Grid~\cite{yu2022monosdf} & QFF-3D~(Ours)
    \end{tabular}
\vspace{-1em}
    \captionof{figure}{We propose Quantized Fourier Features~(QFF), an easy-to-train, memory-efficient yet expressive neural field representation.
    We compare the multilayer perceptron (MLP), multi-resolution grids, and our proposed QFF for neural surface reconstruction over sparse, 3-view reconstructions in the DTU Dataset~\cite{aanaes2016large}.
    The QFF model encodes high-frequency details better than the baseline MLP and preserves smoothness where appropriate, resulting in better geometry compared to the multi-resolution grids. 
    }
    \label{fig:teaser}
\end{center}%
}]

\footnotetext[1]{The work is not related to the author’s position at Amazon.}

\begin{abstract}
Multilayer perceptrons (MLPs) learn high frequencies slowly.  Recent approaches encode features in spatial bins to improve speed of learning details, but at the cost of larger model size and loss of continuity. Instead, we propose to encode features in bins of Fourier features that are commonly used for positional encoding.  We call these Quantized Fourier Features (QFF). As a naturally multiresolution and periodic representation, our experiments show that using QFF can result in smaller model size, faster training,  and better quality outputs for several applications, including Neural Image Representations (NIR), Neural Radiance Field (NeRF) and Signed Distance Function (SDF) modeling.  QFF are easy to code, fast to compute, and serve as a simple drop-in addition to many neural field representations. 

\end{abstract}


\section{Introduction}

Deep networks' impressive ability as function approximators has been demonstrated, \eg for physics~\cite{raissi2019physics}, compression~\cite{strumpler2021implicit}, and 3D modeling~\cite{mildenhall2021nerf,park2019deepsdf}. Yet, learning high-frequency functions is still a challenge. Deep networks tend to learn low-frequency signals first, and higher frequencies later, requiring long training times. This phenomenon, known as spectral bias~\cite{rahaman2019spectral} or frequency principle~(\textit{f-principle})~\cite{xu2019frequency}, is universally observed in multilayer perceptron (MLP) architectures. For instance, optimizing Neural Radiance Field (NeRF)~\cite{mildenhall2021nerf} MLP models takes hours, or even days for large scenes~\cite{mildenhall2021nerf}. The success of NeRF relies on encoding position with multi-frequency sinusoidal coefficients~\cite{mildenhall2021nerf,tancik2020fourfeat} that are amenable to linear functions, but, while quality greatly improves, convergence is still slow because changes to individual parameters can have widespread effects.  Much faster convergence can be achieved by partitioning the function's domain into local components, \eg voxels in a 3D scene, with parameters that can be independently optimized~\cite{muller2022instant,reiser2021kilonerf}. However, spatial gridding reduces continuity and requires multi-resolution grids with many parameters. 

Our \textbf{key insight} is that quantizing the multi-frequency sinusoidal coefficients provides a naturally multiresolution representation that enables the high quality of MLP models and fast training of spatial quantization approaches, with minimal computational and memory cost. A feature vector is learned for each quantized value of a coefficient, and these features are added to the continuous coefficients as the input to an MLP. We call these quantized Fourier features (QFF). By quantizing multiple frequencies, the QFF maintain continuity and high spatial resolution without redundancy.  When modeling functions with sparse outputs, such as occupancy in 3D scenes, the periodicity of QFF provides further advantage of encoding the full domain at high resolution without using many parameters to encode completely empty space. QFF can be computed per input dimension ({\em QFF-Lite}), or factorized and composed into multiple 2D and 1D grids ({\em QFF-3D}) to encode more detail and further speed convergence. QFF is simple to code, fast to compute, and easily inserted into many neural field frameworks. Our experiments demonstrate advantages in convergence and quality of models in applications of image encoding, novel view synthesis, and 3D surface modeling.  

In summary, our \textbf{contribution} is a quantized Fourier feature (QFF) that is easy to compute and include in neural representations and provides advantages of small model size, high resolution, fast training, and excellent model quality across several applications.

\section{Related Work}
Neural networks tend to learn low frequencies of the objective function before high frequencies. This phenomenon is known as spectral bias~\cite{rahaman2019spectral} or the Frequency Principle~\cite{xu2019frequency}, and has been shown to hold for synthetic and real data, as well as multilayer perceptrons, convolutional networks, and other architectures~\cite{xu2022overview}.  
This has direct implications for learning Neural Field Representations (NFR), manifested in a variety of applications including physics-inspired neural networks~\cite{raissi2019physics,brunton2020machine}, shape fitting~\cite{park2019deepsdf,chabra2020deepls}, image compression ~\cite{sitzmann2020implicit}, and neural radiance fields ~\cite{mildenhall2021nerf,barron2021mip,martinbrualla2020nerfw,srinivasan2021nerv}. NFR require longer convergence time when learning high-dimensional, high-frequency functions, which is the main bottleneck for large-scale training. 

Multiple approaches has been proposed to embed feature vectors into explicit geometric structures (\eg voxels)~\cite{reiser2021kilonerf,yu2021plenoctrees,liu2020neural,martel2021acorn,muller2022instant}. 
This strategy typically achieve faster convergence (compared to MLP), they inherently use more memory~\cite{muller2022instant} and are known to overfit~\cite{wizadwongsa2021nex,wu2021diver}. A common strategy to avoid overfitting is to employ gradient-based regularization~\cite{fridovich2022plenoxels,gropp2020implicit} to simulate smoothness. To reduce memory for storing explicit mappings, sparse-grid~\cite{liu2020neural} and octree~\cite{takikawa2021neural} are proposed to allocate more parameters near occupied scene regions. 
Spatial Hashing~\cite{muller2022instant} uses a multidimensional hash table to map input coordinates to random hash table indices. This saves storage but creates a discontinuous representation, which leads to artifacts in inherently smooth functions such as signed distance fields~(SDF)~\cite{yu2022monosdf}.
Approaches such as TensoRF~\cite{chen2022tensorf} reduce parameters by decomposing multi-dimensional arrays into compositions of lower-dimensional arrays.  
Many of these approaches require knowledge of a bounding volume or occupied portions of the scene, necessitating a coarse-to-fine training strategy is used to first identifying volume of interest in coarse volume, and crop-and-resize to fit the finer scenes. 

Fig.~\ref{fig:representation} compares several representations. Our QFF representation uses explicit geometric structure, but in the form of quantized Fourier components so that it is continuous, compact, and multiscale, and does not require prior knowledge of the bounding box or occupied space. We represent 3D volumes, either factorized per dimension ({\em QFF-Lite}) or by composing 2D arrays ({\em QFF-3D}), similar to TensoRF. Further, the QFF representation is based on commonly used positional encodings and easily inserted into most Neural Field Representations.   
\begin{table}[]
    \centering
    \captionsetup{type=figure}
    \begin{tabular}{cl}
 \raisebox{2.2\normalbaselineskip}[0pt][0pt]{{\parbox{1.1cm}{\centering\footnotesize Positional \\ Encoding \\ \cite{mildenhall2021nerf, tancik2020fourfeat}}}} 
        \includegraphics[width=.9\linewidth,keepaspectratio]{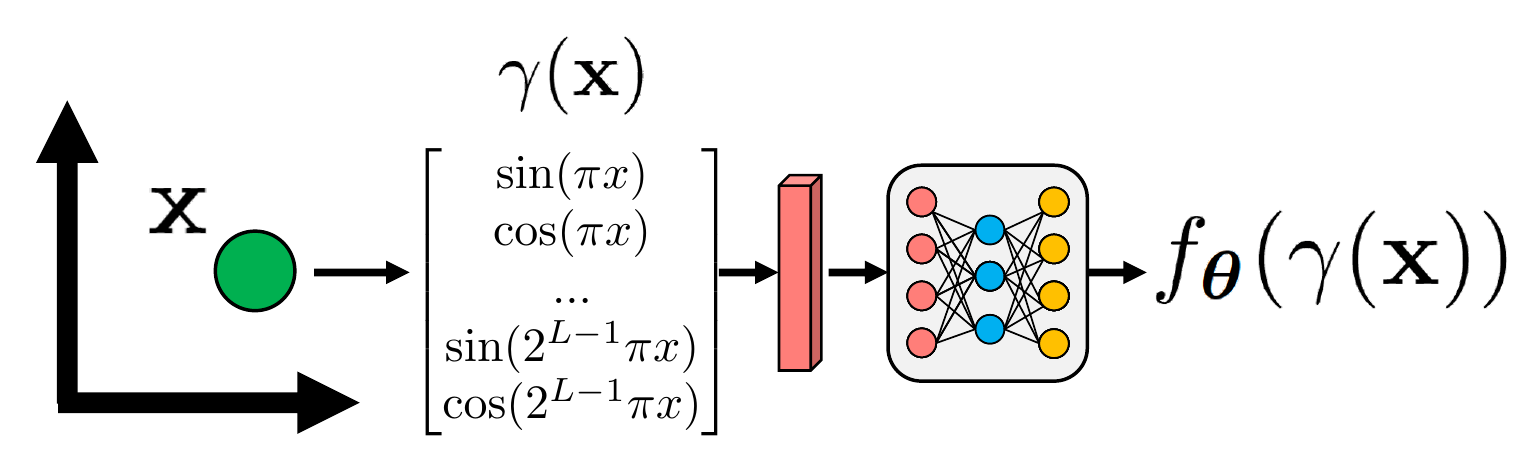}\\
 \raisebox{2.2\normalbaselineskip}[0pt][0pt]{{\parbox{1.1cm}{\centering\footnotesize Explicit \\ Grid \cite{fridovich2022plenoxels, wu2021diver, muller2022instant}}}} 
        \includegraphics[width=.9\linewidth,keepaspectratio]{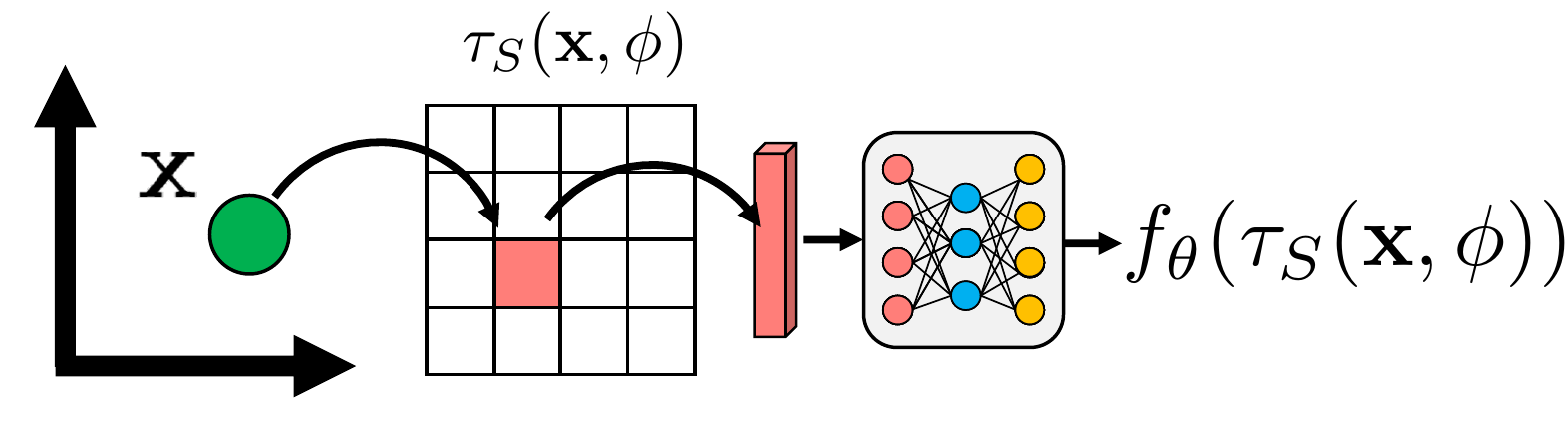}\\
 \raisebox{2.2\normalbaselineskip}[0pt][0pt]{{\parbox{1.1cm}{\centering\footnotesize Tensorial \\ Grid \\ \cite{chen2022tensorf, chan2022efficient}}}} 
        \includegraphics[width=.9\linewidth,keepaspectratio]{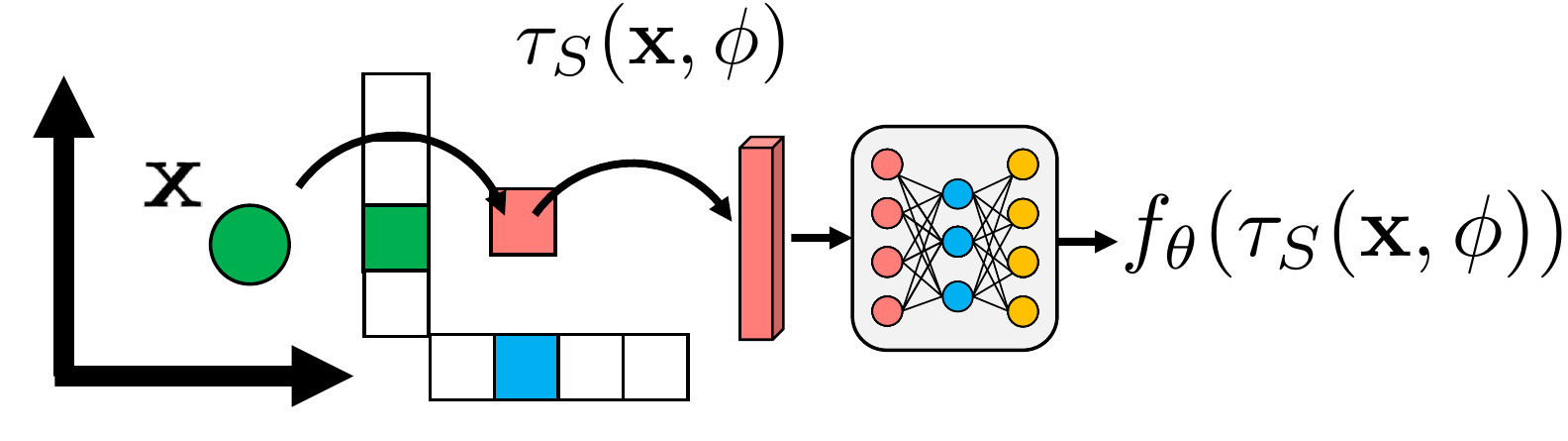}\\
 \raisebox{2.2\normalbaselineskip}[0pt][0pt]{{\parbox{1.1cm}{\centering\footnotesize Quantized \\ Fourier \\ (Ours)}}} 
        \includegraphics[width=.9\linewidth,keepaspectratio]{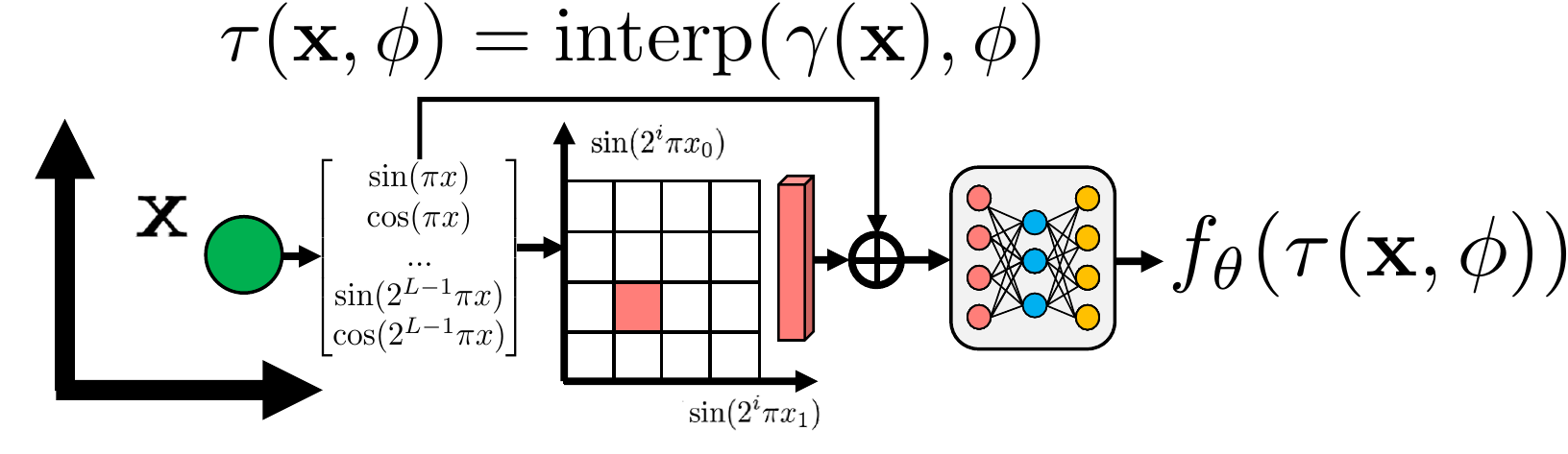}
    \end{tabular}
\captionof{figure}{\textbf{Comparison between different neural field representations.} From top to bottom: positional encoding, explicit spatial grids, tensorial grids, and our proposed quantized Fourier features.}
\label{fig:representation}
\end{table}

\label{fig:overview}
\section{Background: Neural Field Representations}

\subsection{Neural Fields with Positional Encodings} 
Neural field representations approximate the mapping from coordinates $\bx \in \bbR^K$ to signal values $\bv \in \bbR^D$ (e.g., color or opacity) with a learnable neural network $f_{\btheta}(\cdot)$, often instantiated with a multilayer perceptron (MLP). An essential recipe for neural fields is positional encoding~\cite{mildenhall2021nerf, tancik2020fourfeat}, which maps the Euclidean coordinates input to sinusoidal activations across $L$ different frequency levels: 
\begin{equation}
\label{eq:pos_enc}
 \gamma(x_k) = [\sin(2^{0\ldots L-1}\pi x_k), \cos(2^{0\ldots L-1}\pi x_k)] 
\end{equation}
$\gamma(\bx) = \text{concat}([\gamma(x_0)...\gamma(x_{K-1})])$ is a continuous, multiscale, periodic representation of $\bx$ along each coordinate.  Then, $\bv$ is computed as:
\begin{equation}
\bv = f_{\btheta}(\gamma(\bx))
\end{equation}
Neural fields representations with positional encodings can capture high-frequency structures in the scene, making them suitable for representing complex spatial-temporal signals across various modalities, such as images~\cite{sitzmann2020implicit}, videos~\cite{pumarola2021d}, and 3D shapes~\cite{wang2021neus,mildenhall2021nerf}. However, it takes a long time to converge, because all MLP parameters have to update for every pair of inputs and outputs.

\begin{figure}
    \centering
    \includegraphics[width=0.48\textwidth]{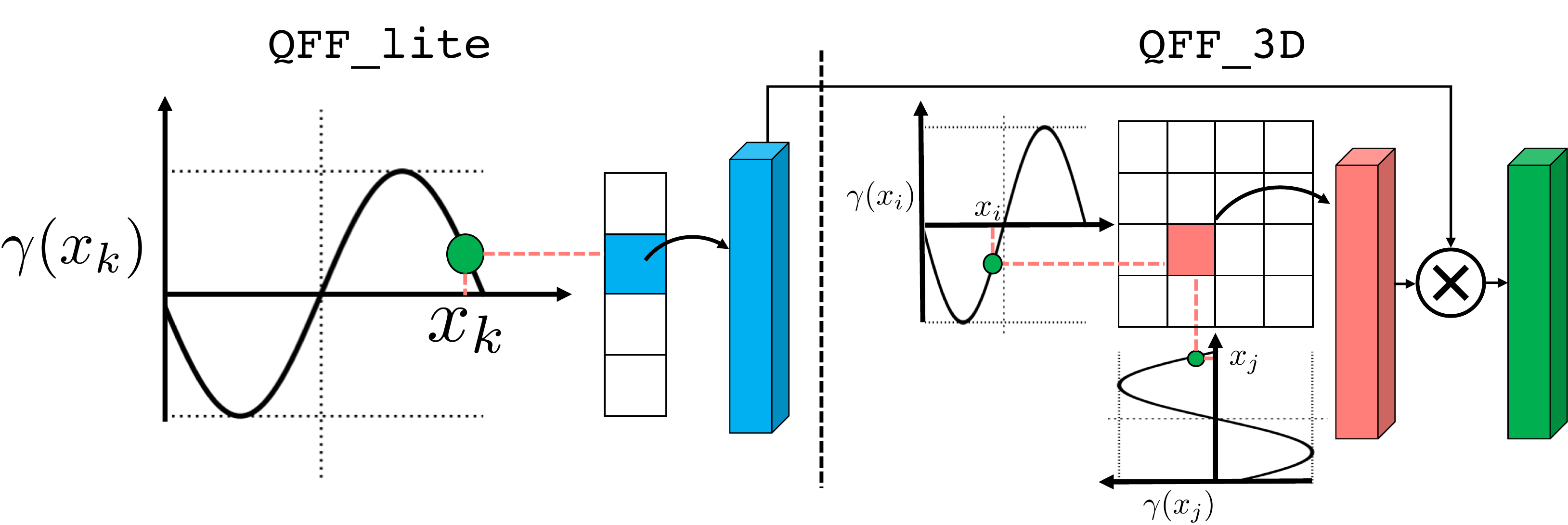}
    \caption{
 Visualization of single-dimensional $\texttt{QFF-lite}$ and $\texttt{QFF-3D}$ feature sampling procedure (applied separately for each positional encoding component). 
 $\texttt{QFF-lite}$: Each positional encoding $\gamma(x_k)$ is discretized into $M$ bins. Each bin contains a multi-dimensional learnable feature $\phi$. The encoded value $\gamma(x_k))$ is used as an index to query features from $\phi$: $\tau(\gamma(x_k), \phi)$.
 $\texttt{QFF-3D}$: We query $\texttt{QFF-lite}$ features $\tau(\gamma(x_k), \phi)$. We then build $M\times M$ bins and use the positional encoding of $(\gamma(x_i), \gamma(x_j))$ to query the corresponding features using bilinear interpolation. We compute the element-wise product between $\tau(x_i, x_j, \phi)$ and $\tau(\gamma(x_k), \phi)$ to get the final feature. 
    }
    \label{fig:arch_1d}
\end{figure}

\subsection{Neural Fields with Explicit Features} 
An alternative paradigm~\cite{chen2022tensorf,peng2020convolutional,muller2022instant,wu2021diver, fridovich2022plenoxels,zhang2022nerfusion} is to explicitly encode features at the input position $\tau_s(\bx; \bphi)$, where $\bphi$ are learnable feature tensors that are linearly interpolated based on $\bx$ to get feature values.
Often, the feature values are added to the positional encoding values as input to the network, which produces $\bv$.
Popular choices for $\tau_s(\bx; \bphi)$ include trilinearly interpolated voxel features~\cite{wu2021diver}; tri-plane representation~\cite{chen2022tensorf,peng2020convolutional} and spatial voxel hashing~\cite{muller2022instant}. Compared to the positional encoding-based neural field representation, such representations are significantly faster to train and evaluate. Furthermore, the neural fields are readily compositional and scalable. But because existing explicit feature representations are in the {\it spatial} domain, they are sensitive to resolution and may waste memory representing features in empty portions of space.


\section{Quantized Fourier Features (QFF)}

We present Quantized Fourier Features~(QFF), an easy-to-train, memory-efficient yet expressive neural field representation. 
Our proposed approach combines the best worlds of explicit feature representation and frequency-based positional encoding. The key idea is to store explicit features in the Fourier domain and use quantized positional encoding to query the feature. 

\begin{figure}
    \centering
    \includegraphics[width=0.45\textwidth]{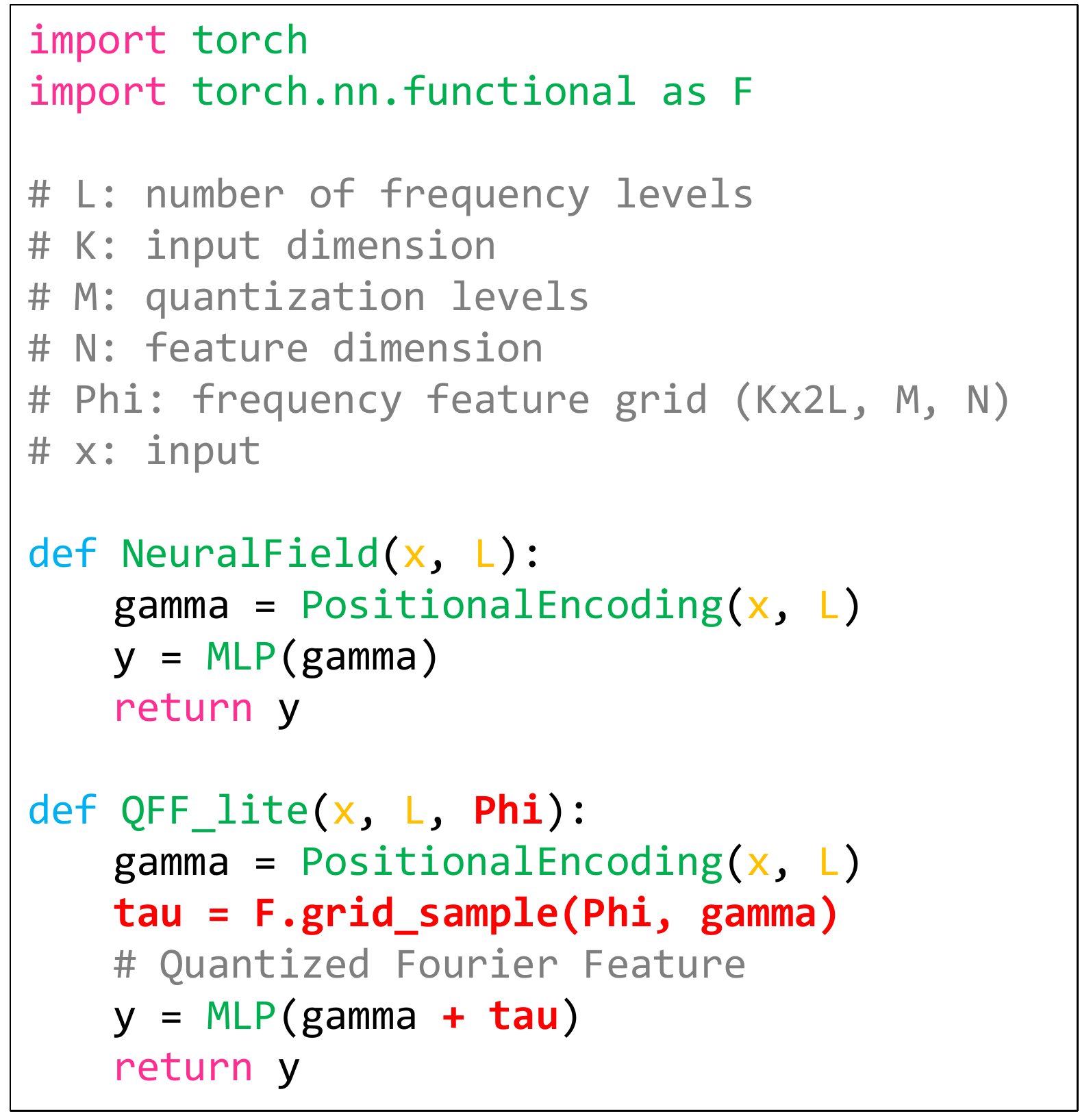}
    \caption{
    \textbf{Apply \textit{QFF} to existing methods.} We show PyTorch~\cite{paszke2019pytorch} pseudo-code to apply QFF to existing MLP based systems, which can be implemented in less than 4 lines of code. 
    }
    \label{alg:apply}
\end{figure}

\subsection{QFF-Light} 
We first explain QFF when encoding each input dimension separately.  
As shown in Figure~\ref{fig:arch_1d}, we create $M$ bins for each $i$-th position encoding component from Eq.~\ref{eq:pos_enc}, and each bin stores a learnable $N$-dimensional feature vector. This gives us a ${K \times 2L\times M \times N}$-dimensional tensor $\bphi$, where $K$ is the input dimension (e.g., 3 for a volume), $L$ is the number of frequencies, $M$ is the quantization level, and $N$ is feature dimension.
Given the spatial input coordinate value $x_k$, we compute features for each  $i$-th positional encoding value by linear interpolation: 
\begin{equation}
    \tau(x_k; \bphi)_i = \text{interp}(\gamma(x_k)_i, \bphi_{ki})
\end{equation}
where $\gamma(x_k)_i$ denotes the $i$-th positional encoding value of $x_k$ and $\bphi_{ki}$ is a $M \times N$ slice of $\bphi$ after indexing with $k$ and $i$. 
Figure~\ref{fig:arch_1d} depicts the feature computation process.

As QFF and Multi-dimensional QFF are piecewise-linear functions, the derivative at the border of quantization in QFF changes sharply due to the discretization. 
Hence, we add the original positional encoding values to the quantized features to induce smoothness. We name the per-dimension QFF, which adds the positional encoding, as QFF-Lite:
\begin{equation}
    \texttt{QFF-Lite}(x_k; \bphi)_i = \tau(x_k, \bphi)_i + \gamma(x_k)_i.
\end{equation}
The resulting values are concatenated across $k$ and $i$, input to the MLP, and trained via gradient descent. 

\subsection{QFF-3D}
\label{sec:kdqff}

So far, QFF is computed independently for each coordinate. When applying the $1$-D QFF embeddings to multi-dimensional input ($n>1$), the MLPs must learn to model correlations across multi-dimensional features. This brings an additional burden to MLPs and makes them less capable or slower to converge. We can instead store features along a grid or volume. For example, for $2$-D QFF, we create $M\times M$ bins for each positional encoding component across both dimensions, so that $\bphi$ is ${K \times 2L\times M^2 \times N}$-dimensional, and bilinearly interpolate at query time. 

For 3-dimensional data, we use the multi-dimensional QFF by TensoRF~\cite{chen2022tensorf} style of spatial decomposition. In particular, we follow vector-matrix~(VM) decomposition and define QFF-3D as:
\begin{equation}
    \texttt{QFF-3D}(x_0; x_1, x_2,\bphi)_i = \tau(x_0, \bphi)_i \cdot \tau(x_1, x_2; \bphi)_i+\gamma(x_0)_i
    \label{eq:qff-vm}
\end{equation}
Equation~\ref{eq:qff-vm} describes the VM decomposition in $x_0$ dimension. We apply the VM decomposition in all 3 dimensions, as done in TensoRF~\cite{chen2022tensorf}.  

\subsection{Application of QFF}

QFF is fast to compute and differentiable. Unlike spatial gridding approaches, where 3D positions and features directly correspond, each spatial position in QFF is represented in multiple frequencies, and each frequency component is represented at multiple positions, as depicted in Figure~\ref{fig:arch_1d}.  This enables the network to use the parameters to model occupied portions of the scene, without knowing which portions are occupied in advance, improving convergence and resolution for a given number of parameters. 

QFF is also easy to plug and play into the existing neural fields that use Fourier positional encodings.  The addition of QFF is independent of the flow of existing methods, and the computation of positional encodings can be reused when computing the QFF feature.  Figure~\ref{alg:apply} illustrates how to apply QFF to existing methods in few lines of code to speed up training and improve performance.

\section{Experiments}
We experiment with our proposed QFF on a wide range of neural field representations: 
\begin{enumerate}[label={(\arabic*)}]
\item \textbf{Neural Image Representations}~(Section \ref{sec:image}); 
\item \textbf{Neural Radiance Fields}~(Section \ref{sec:nerf});
\item \textbf{Neural Signed Distance Fields}, including few-shot object reconstruction and large-scale scene reconstruction~(Section \ref{sec:nsdf}). 
\end{enumerate}
Through experiments on (1) and (2), we show our improvement over standard positional encoding and efficiency in spatial representation due to spatial hashing. We further demonstrate our ability to learn continuous representations in (3). 
We apply QFF-3D and QFF-Lite to existing state-of-the-art and baseline architectures, to demonstrate the improvement purely due to adding QFF to the representation.
\begin{table}[t]
\centering
\small
\begin{tabular}{l| c c c c}
\toprule
Method                              & Pos. Enc.  & QFF-Lite & Natural & Text\\
\midrule
\multirow{3}{*}{ReLU} 
& $\times$      & $\times$      & 17.74             & 18.38\\
& $\checkmark$  & $\times$      & 29.41             & 30.60\\
& $\checkmark$  & $\checkmark$  & \textbf{30.30}    & \textbf{32.05}\\
\midrule
\multirow{3}{*}{SIREN~\cite{sitzmann2020implicit}} 
& $\times$      & $\times$      & 28.18             & 30.94\\
& $\checkmark$  & $\times$      & 37.06             & \textbf{47.34} \\
& $\checkmark$  & $\checkmark$  & \textbf{37.68}    & 46.92\\
\bottomrule
\end{tabular}

\caption{
    \textbf{Evaluation on supervised Neural Image Representation in PSNR.} The best performing method for each dataset (Netural, Text) for each activation function (ReLU, SIREN) is marked in bold. 
    Overall, we see improvements by applying \textit{QFF-Lite}.
}
\label{tab:megapixel}
\end{table}

\begin{table*}[h]
\centering

\resizebox{\textwidth}{!}{
\begin{tabular}{c|l|cccccccc|c|c|c}
\toprule
\multicolumn{2}{c}{Method}                &   Chair   & Drums   & Ficus   & Hotdog    & Lego  & Materials & Mic   & Ship  & Mean & Params & Steps \\
\midrule
\multirow{4}{*}{Decomp.}
&NeRF~\cite{mildenhall2021nerf}      &   33.00          & 25.01          & 30.13          & 36.18          & 32.54          & 29.62          & 32.91          & 28.65           & 31.00          & 1.19 M         & 100+K\\
&TensoRF-CP~\cite{chen2022tensorf}   &   \textbf{33.60} & 25.17          & 30.72          & \textbf{36.24} & 34.05          & 30.10          & 33.77          & \textbf{28.84}  & 31.56          & 728 K          & \textbf{30 K}\\
&NeRFAcc~\cite{li2022nerfacc}        &   33.32          & \textbf{25.39} & \textbf{32.52} & 35.80          & 33.69          & 29.73          & 33.76          & 28.18           & 31.55          & 618 K          & 50 K\\
&NeRFAcc~(QFF-Lite)                  &   33.36          & 25.33          & 31.97          & 35.70          & \textbf{34.16} & \textbf{30.15} & \textbf{33.90} & 28.20           & \textbf{31.59} & \textbf{522 K} & 50 K\\
\midrule
\multirow{3}{*}{Comp.}
&Instant-NGP~\cite{muller2022instant}&   35.00          & \textbf{26.02} & 33.51          & 37.40          & 36.39          & 29.78          & 36.22          & \textbf{31.10} & 33.18          & 12.6 M & 50 K\\
&TensoRF-VM~\cite{chen2022tensorf}   &   \textbf{35.76} & 26.01          & \textbf{33.99} & \textbf{37.41} & 36.46          & \textbf{30.12} & 34.61          & 30.77          & 33.14          & 17.6 M & \textbf{30 K}\\
&NeRFAcc~(QFF-3D)    &   35.69          & 25.97          & 33.70          & 37.07          & \textbf{36.68} & 30.07          & \textbf{37.53} & 29.97          & \textbf{33.35} & \textbf{9.82 M} & 50 K\\
\bottomrule
\end{tabular}
}

\caption{
    \textbf{PSNR evaluation on test images of the NeRF Synthetic dataset}. 
    The top and bottom rows show methods without and with feature composition, respectively.
    For the baseline methods, we use the numbers reported in original papers. 
    We mark the best method for each section in bold. Our method achieves better mean image quality for novel view synthesis with fewer model parameters.
}
\label{tab:nerf}
\end{table*}

\begin{table*}
\footnotesize
\setlength\tabcolsep{.5pt} 
\centering
\begin{tabular}{ccccc}
    \includegraphics[width=0.18\textwidth]{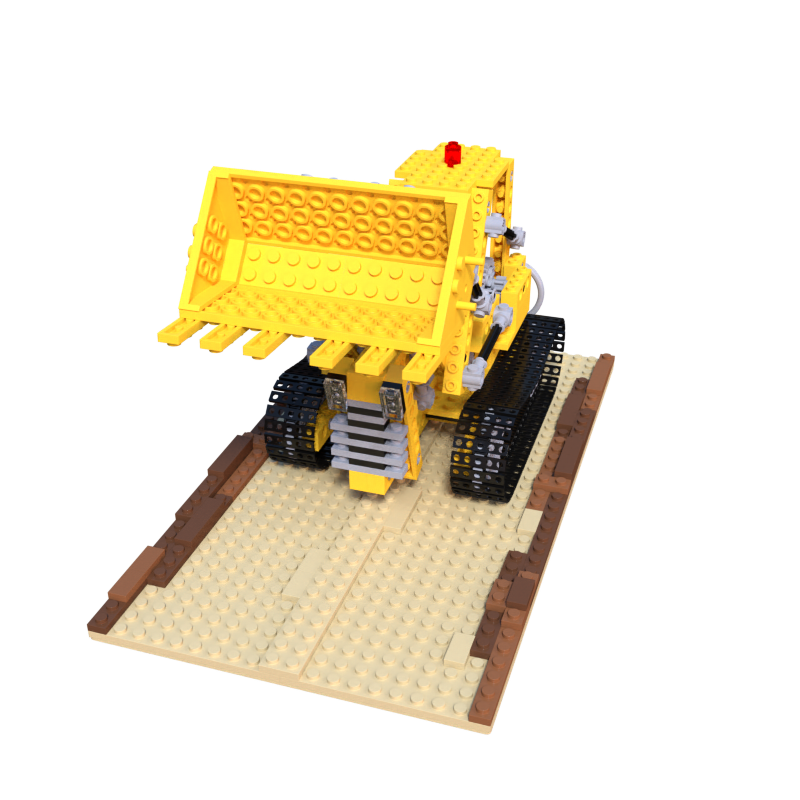} &
    {\tikz{
    \node[inner sep=0pt](gt) at (0, 0) {\includegraphics[trim={7cm 15cm 15cm 7cm},clip,width=0.18\textwidth]{images/nerf_synthetic/lego_155_gt.png}};
    }} & 
    {\tikz{
    \node[inner sep=0pt](tensrf) at (0, 0) {\includegraphics[trim={7cm 15cm 15cm 7cm},clip,width=0.18\textwidth]{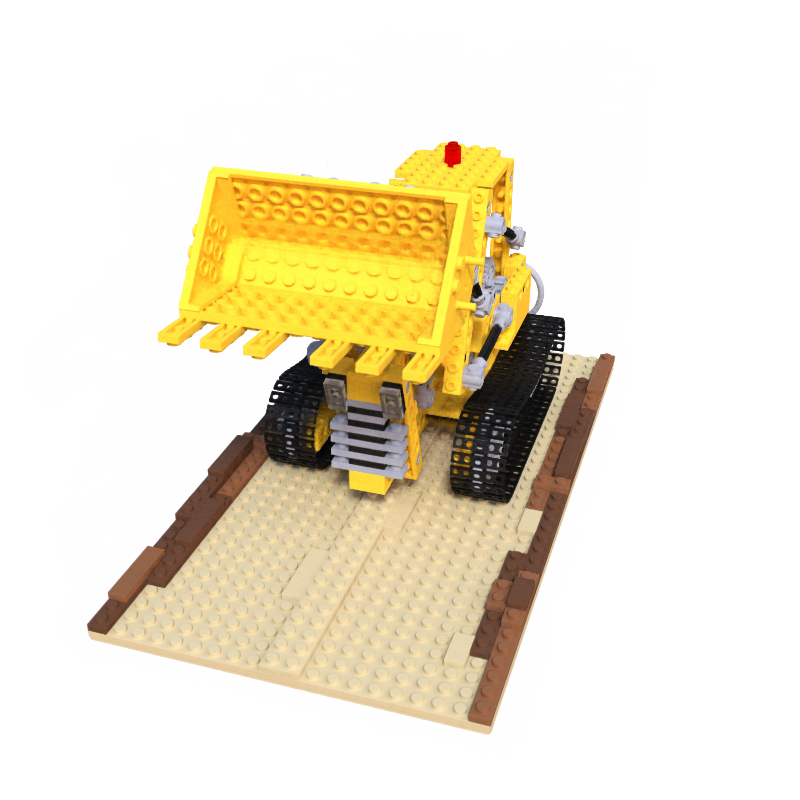}};
    \node[draw=black, draw opacity=1.0, line width=.3mm, fill opacity=0.7,fill=white, text opacity=1] at (-.5, -1.3) {\small PSNR = 34.05};
    }} 
    & 
    {\tikz{
    \node[inner sep=0pt](acc) at (0, 0) {\includegraphics[trim={7cm 15cm 15cm 7cm},clip,width=0.18\textwidth]{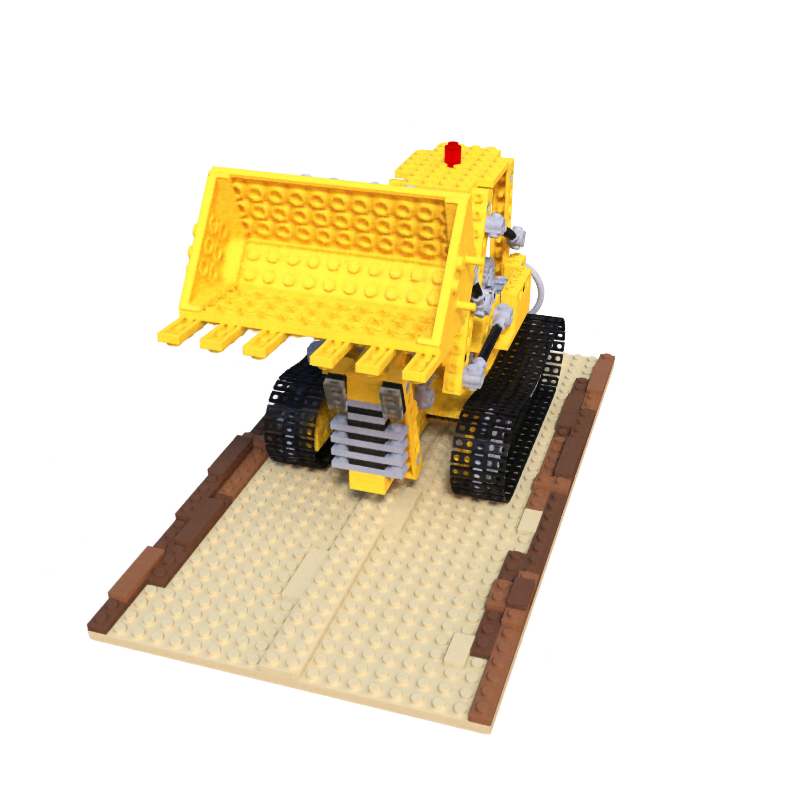}};
    \node[draw=black, draw opacity=1.0, line width=.3mm, fill opacity=0.7,fill=white, text opacity=1] at (-.5, -1.3) {\small PSNR = 33.69};
    }} 
    &
    {\tikz{
    \node[inner sep=0pt](qff) at (0, 0) {\includegraphics[trim={7cm 15cm 15cm 7cm},clip,width=0.18\textwidth]{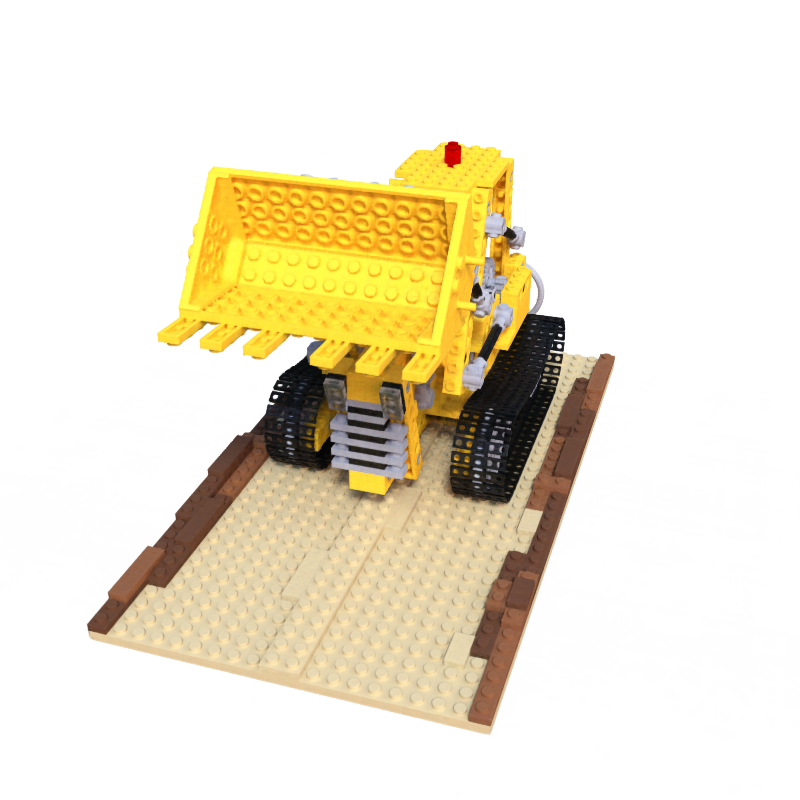}};
    \node[draw=black, draw opacity=1.0, line width=.3mm, fill opacity=0.7,fill=white, text opacity=1] at (-.5, -1.3) {\small PSNR = 34.16};
    }} \\
    \textit{Lego} & Ground Truth & TensoRF-CP~\cite{muller2022instant} & NeRFAcc~\cite{li2022nerfacc} & QFF-Lite~(Ours)\\

    \includegraphics[width=0.18\textwidth]{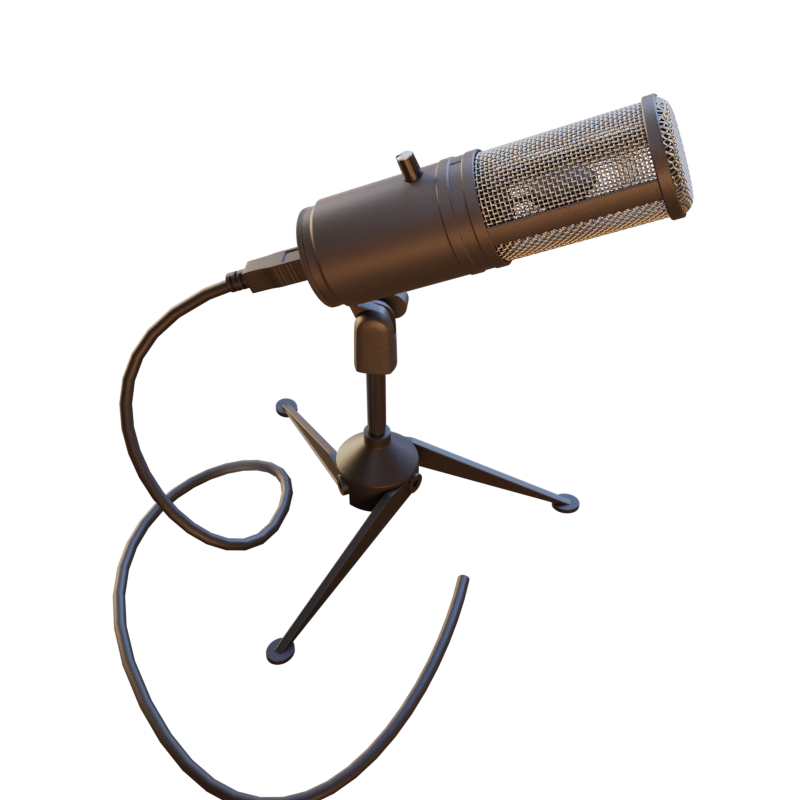} &
    \includegraphics[trim={19cm 19cm 4cm 4cm},clip,width=0.18\textwidth]{images/nerf_synthetic/mic_27_gt.png} &
    {\tikz{
    \node[inner sep=0pt](tensrf) at (0, 0) {\includegraphics[trim={19cm 19cm 4cm 4cm},clip,width=0.18\textwidth]{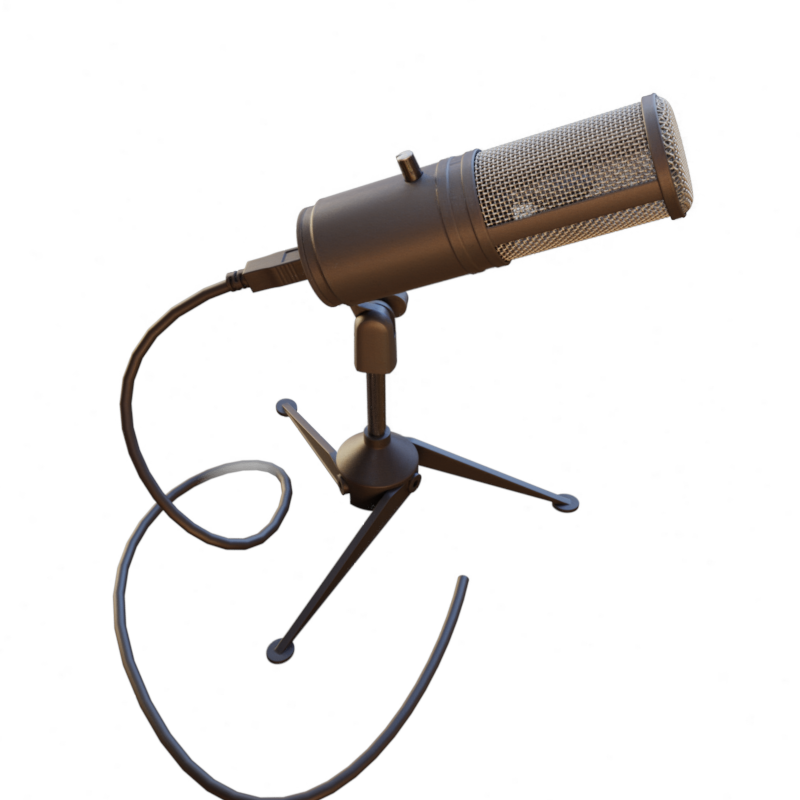}};
    \node[draw=black, draw opacity=1.0, line width=.3mm, fill opacity=0.7,fill=white, text opacity=1] at (-.5, -1.3) {\small PSNR = 36.22};
    }} 
    & 
    {\tikz{
    \node[inner sep=0pt](acc) at (0, 0) {\includegraphics[trim={19cm 19cm 4cm 4cm},clip,width=0.18\textwidth]{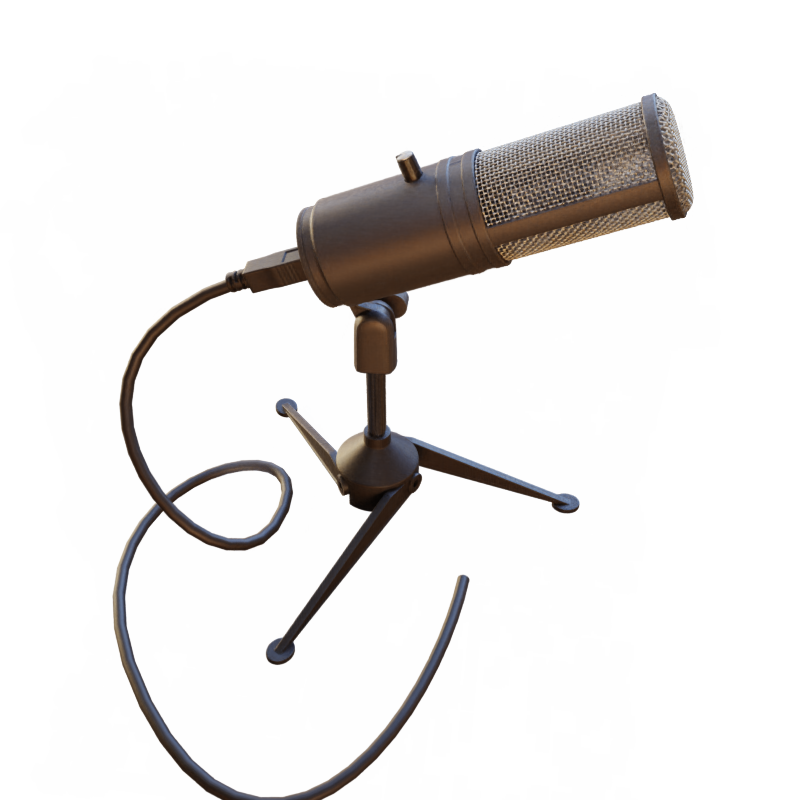}};
    \node[draw=black, draw opacity=1.0, line width=.3mm, fill opacity=0.7,fill=white, text opacity=1] at (-.5, -1.3) {\small PSNR = 34.61};
    }} 
    &
    {\tikz{
    \node[inner sep=0pt](qff) at (0, 0) {\includegraphics[trim={19cm 19cm 4cm 4cm},clip,width=0.18\textwidth]{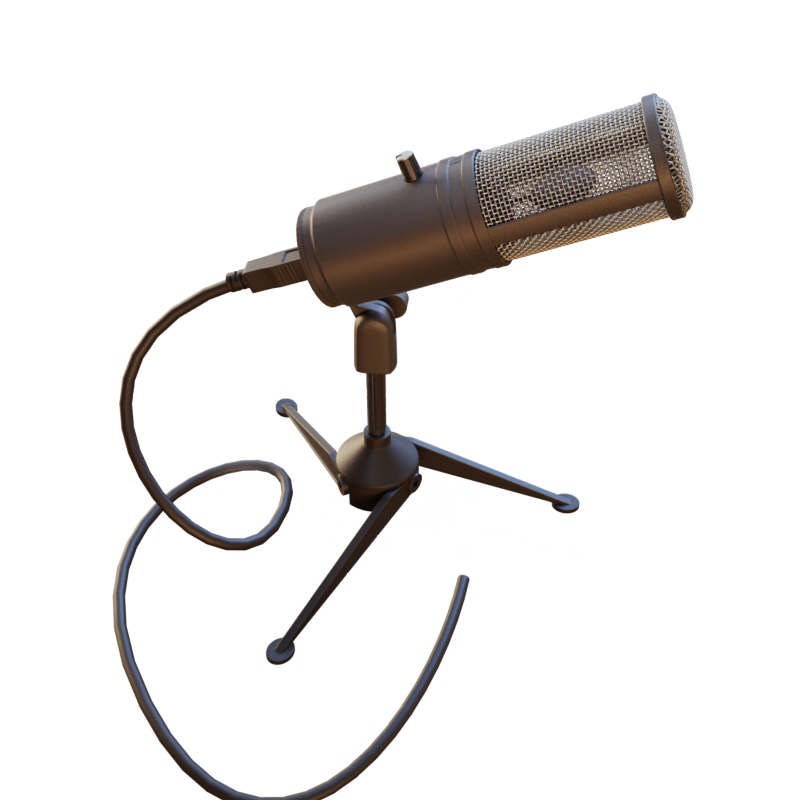}};
    \node[draw=black, draw opacity=1.0, line width=.3mm, fill opacity=0.7,fill=white, text opacity=1] at (-.5, -1.3) {\small PSNR = 37.53};
    }} \\
    \textit{Microphone} & Ground Truth & Instant-NGP~\cite{muller2022instant} & TensoRF-VM~\cite{chen2022tensorf} & QFF-3D~(Ours) \\
    \includegraphics[width=0.18\textwidth]{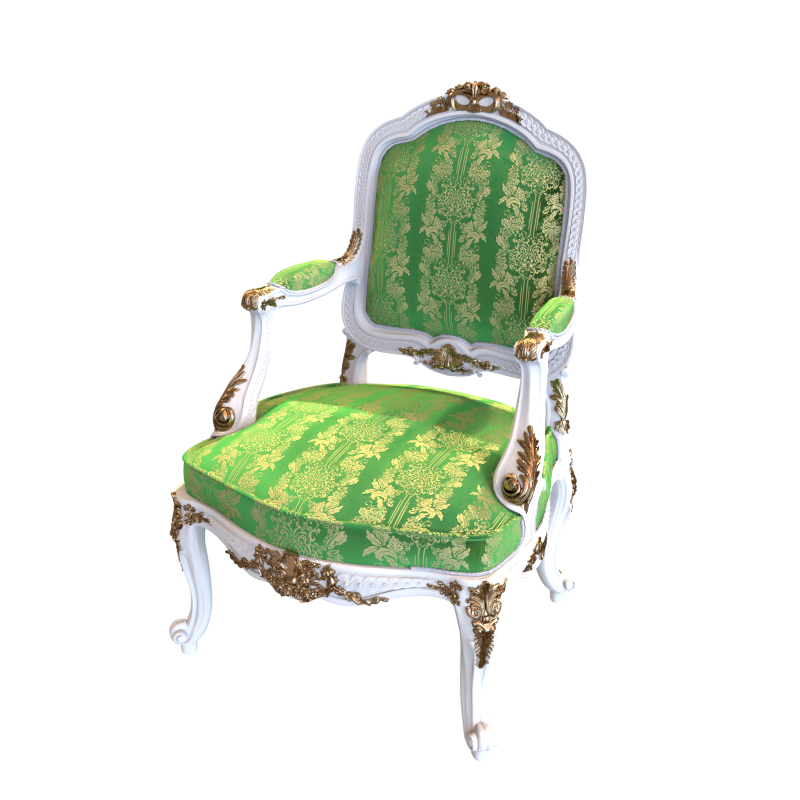} &
    \includegraphics[trim={12cm 10cm 12cm 14cm},clip,width=0.18\textwidth]{images/nerf_synthetic/chair_58_gt.png} &
    {\tikz{
    \node[inner sep=0pt](tensrf) at (0, 0) {\includegraphics[trim={12cm 10cm 12cm 14cm},clip,width=0.18\textwidth]{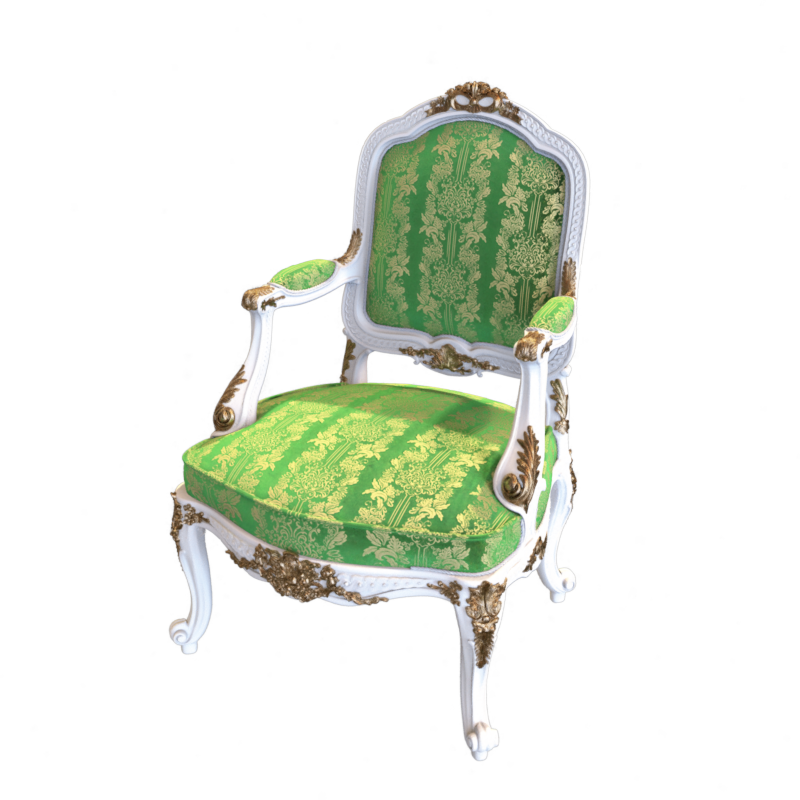}};
    \node[draw=black, draw opacity=1.0, line width=.3mm, fill opacity=0.7,fill=white, text opacity=1] at (-.5, -1.3) {\small PSNR = 35.00};
    }} 
    & 
    {\tikz{
    \node[inner sep=0pt](acc) at (0, 0) {\includegraphics[trim={12cm 10cm 12cm 14cm},clip,width=0.18\textwidth]{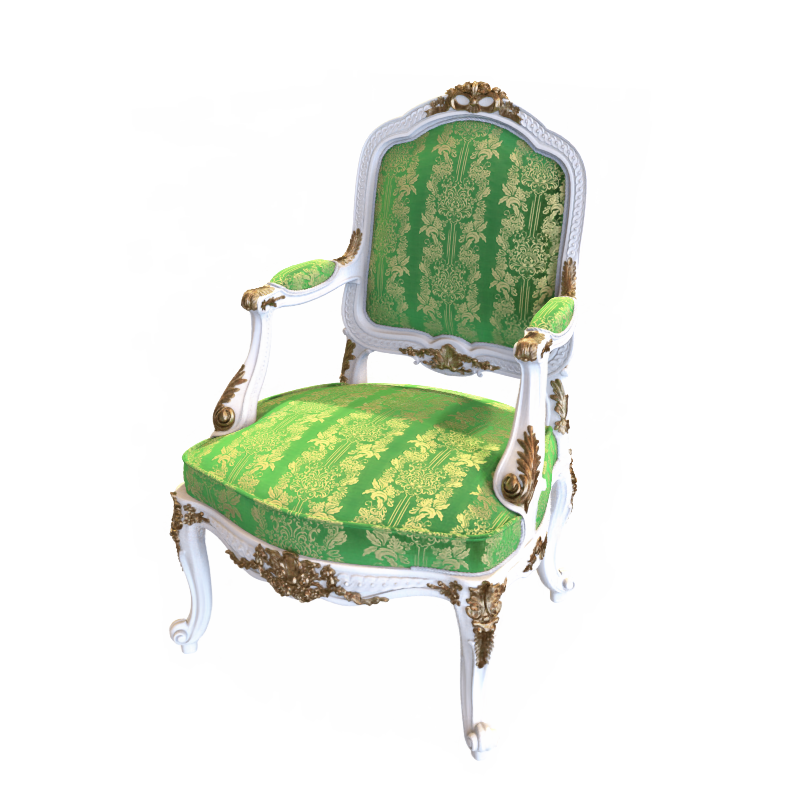}};
    \node[draw=black, draw opacity=1.0, line width=.3mm, fill opacity=0.7,fill=white, text opacity=1] at (-.5, -1.3) {\small PSNR = 35.76};
    }} 
    &
    {\tikz{
    \node[inner sep=0pt](qff) at (0, 0) {\includegraphics[trim={12cm 10cm 12cm 14cm},clip,width=0.18\textwidth]{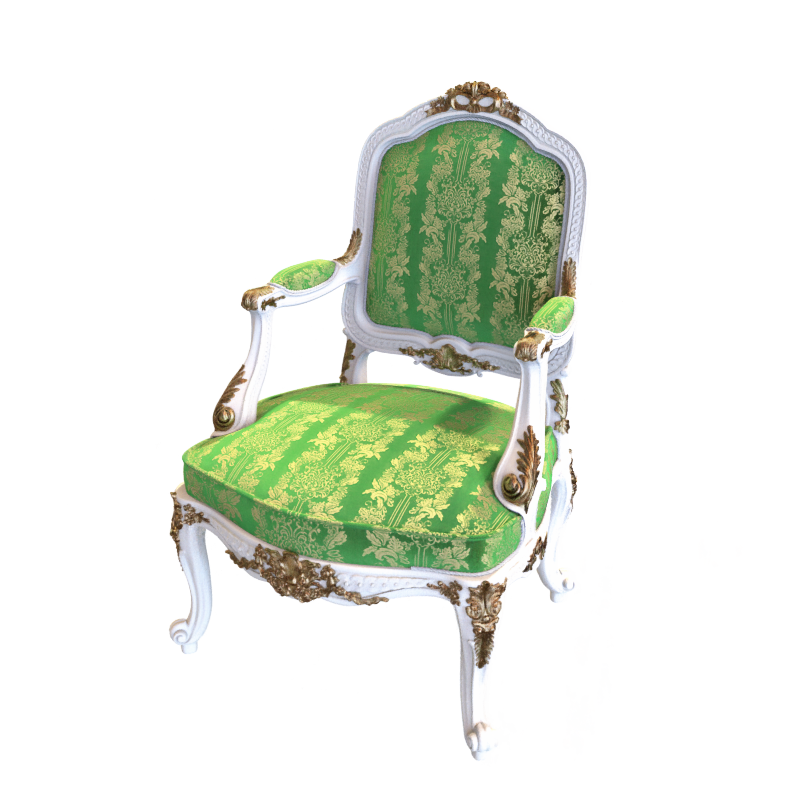}};
    \node[draw=black, draw opacity=1.0, line width=.3mm, fill opacity=0.7,fill=white, text opacity=1] at (-.5, -1.3) {\small PSNR = 35.69};
    }} \\
    \textit{Chair} & Ground Truth & Instant-NGP~\cite{muller2022instant} & TensoRF-VM~\cite{chen2022tensorf} & QFF-3D~(Ours)
\end{tabular}
\captionof{figure}{\textbf{Comparison of different neural radiation field representations.} The first row compares the decomposed representation of the scene \textit{Lego}. 
The bottom two rows compare the composed representations of the scene \textit{Microphone} and \textit{Chair}.
We obtain images of TensoRF~\cite{chen2022tensorf} from the results provided by the authors, and NeRFAcc~\cite{li2022nerfacc} and Instant-NGP~\cite{muller2022instant} results by running the code provided by the authors.
} 
\label{fig:nerf_synthetic}
\end{table*}
\subsection{Neural Image Representations}
\label{sec:image}
\textbf{Dataset.} We use natural and text mega-pixel images from Tancik et al.~\cite{tancik2020fourfeat}. Natural images contain high frequency signals and text images contain sparse signals.

\textbf{Evaluation Metrics.} We evaluate the reconstruction ability by measuring the peak signal-to-noise ratio (PSNR).

\textbf{Implementation.} For 2D image representation, we use QFF-Lite with $N = 1$: a single value is added to each positional encoding value after sampling. We use $L=128$ for the number of scales and $M=2^7$ for the feature bin resolution. We use Adam~\cite{kingma2014adam} optimizer with $lr=5e^{-4}$ for both QFF and MLP parameters. 

\textbf{Results.} We compare application of our QFF-Lite on 
ReLU and SIREN~\cite{sitzmann2020implicit} activation functions, with and without our method for supervised 2D image reconstruction task. Table~\ref{tab:megapixel} shows the results for natural images and text images reconstruction.
For both ReLU and SIREN, after applying QFF-Lite, we achieve a slight improvement (about 0.6) in PSNR on natural images. ReLU using QFF-Lite in text images achieves a significant improvement (1.45).
We see a slight drop in PSNR for text images on SIREN, but we note that the PSNR of SIREN text reconstruction with positional encoding has peaked; there is not much room for improvement. 

\begin{table*}[t!]
\resizebox{\textwidth}{!}{
\centering
\begin{tabular}{ccccc}
    \includegraphics[width=0.19\textwidth]{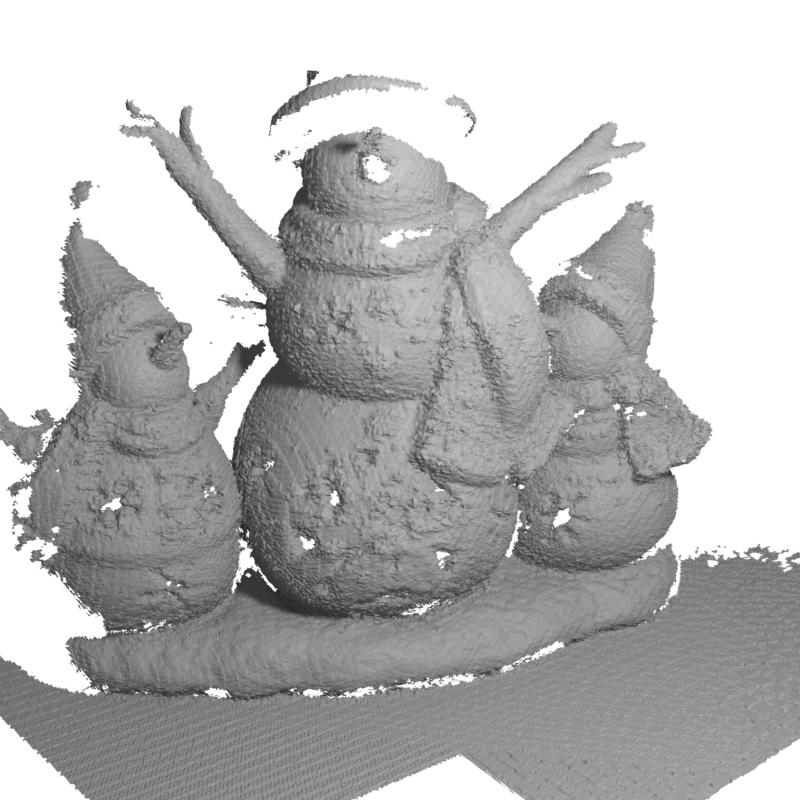} &
    \includegraphics[width=0.19\textwidth]{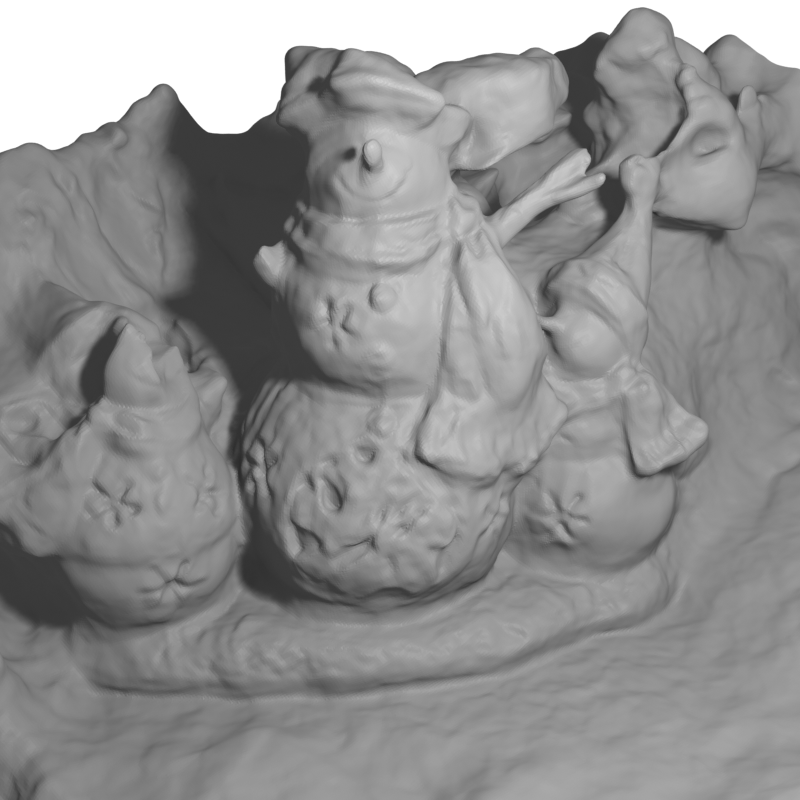} &
    \includegraphics[width=0.19\textwidth]{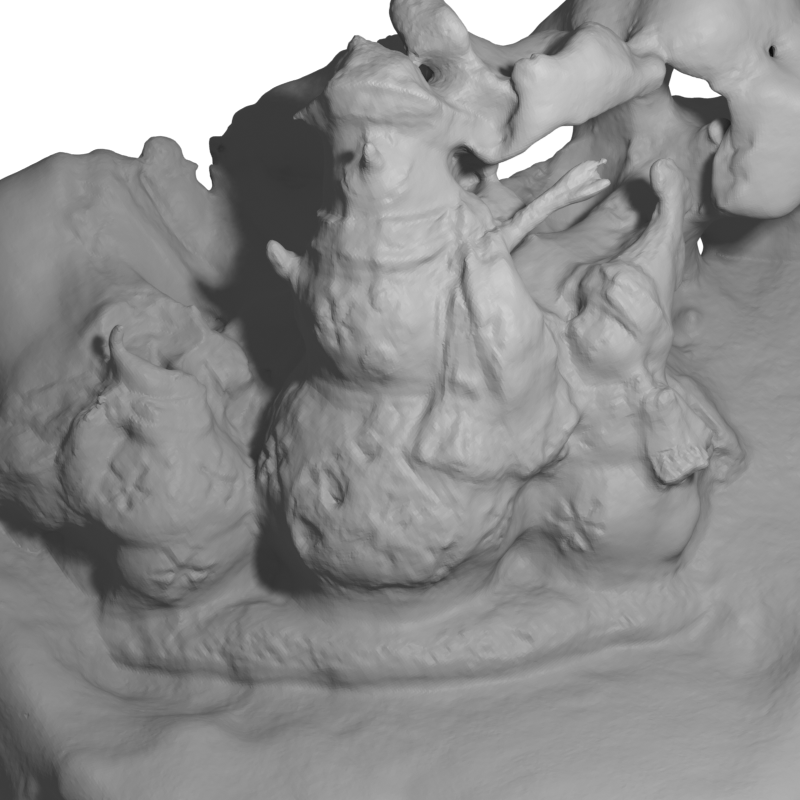} &
    \includegraphics[width=0.19\textwidth]{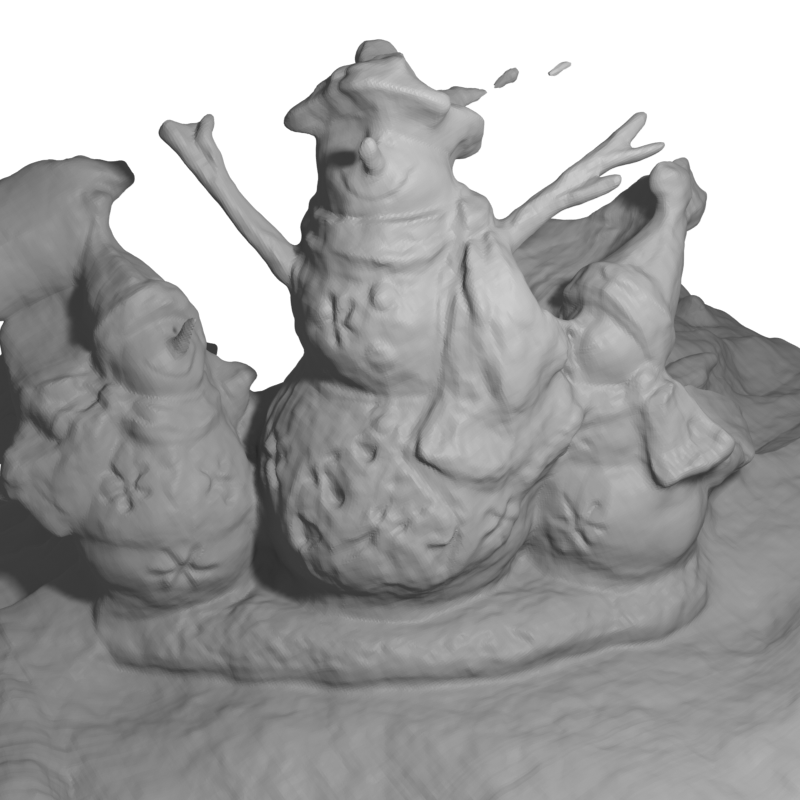} &
    \includegraphics[width=0.19\textwidth]{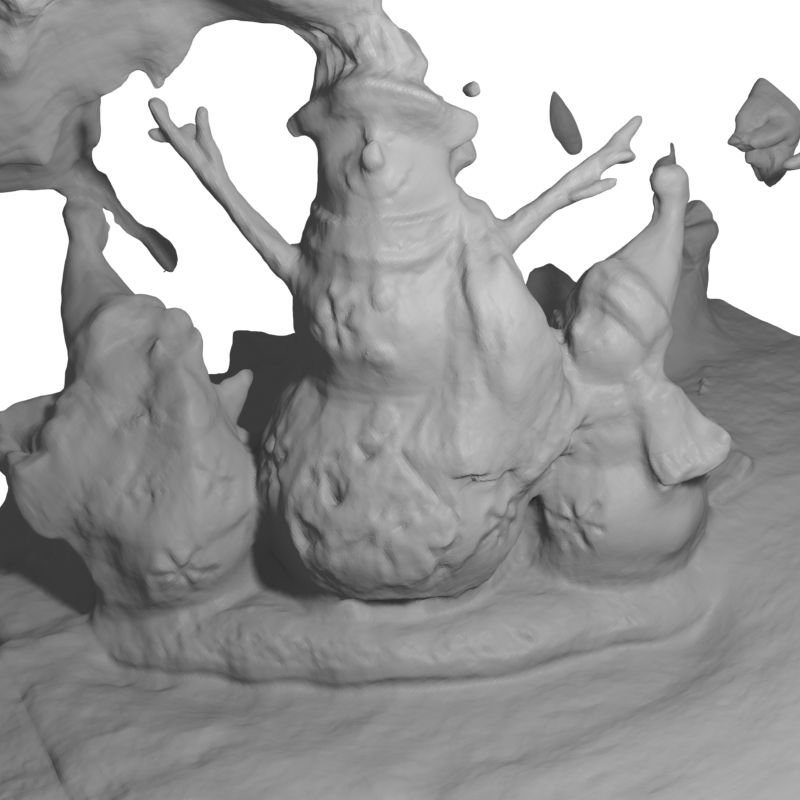}\\
    \includegraphics[width=0.19\textwidth]{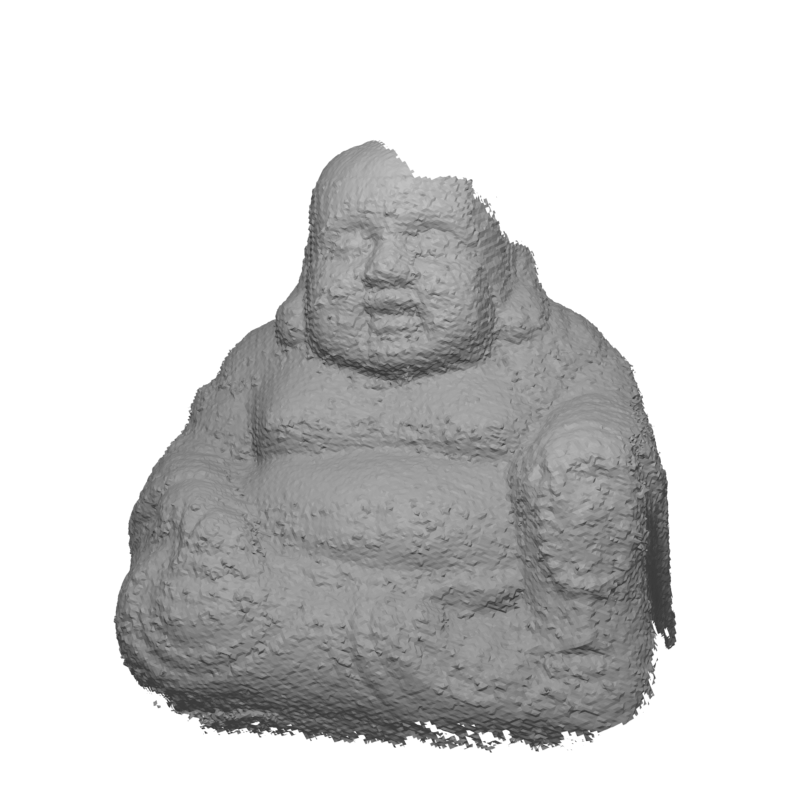} &
    \includegraphics[width=0.19\textwidth]{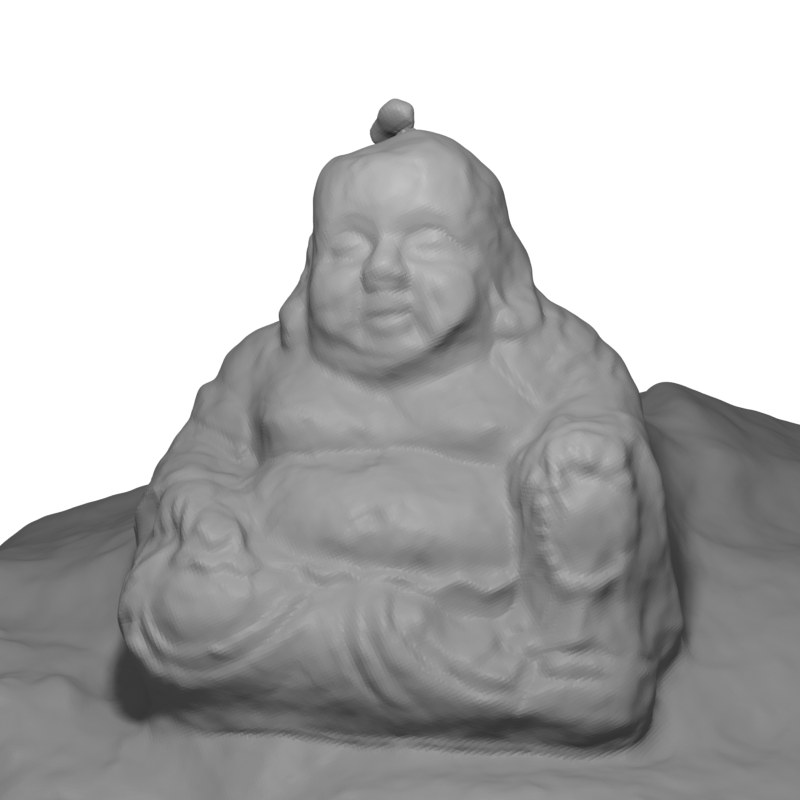} &
    \includegraphics[width=0.19\textwidth]{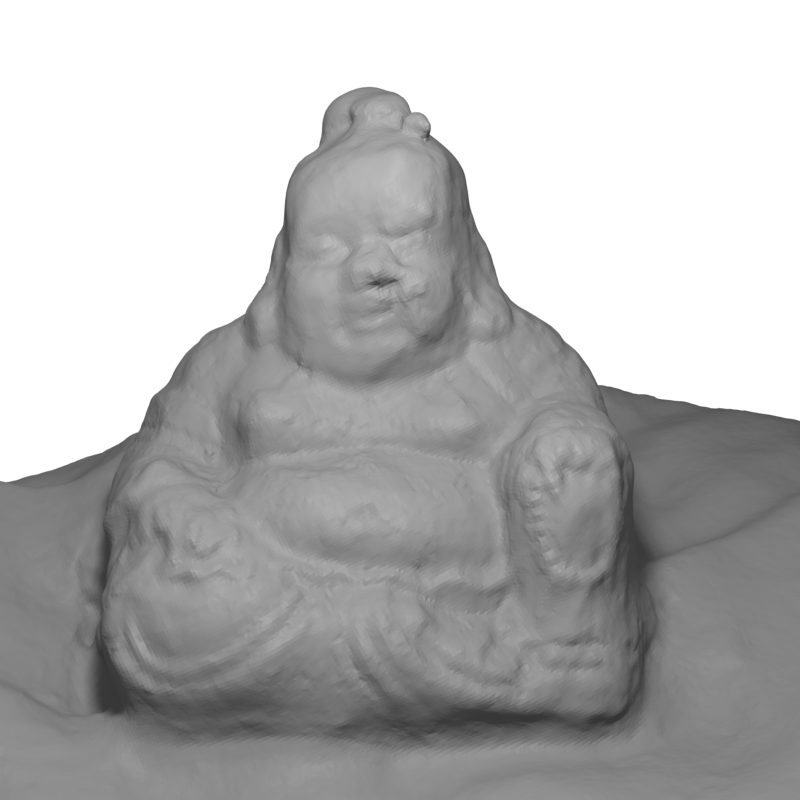} &
    \includegraphics[width=0.19\textwidth]{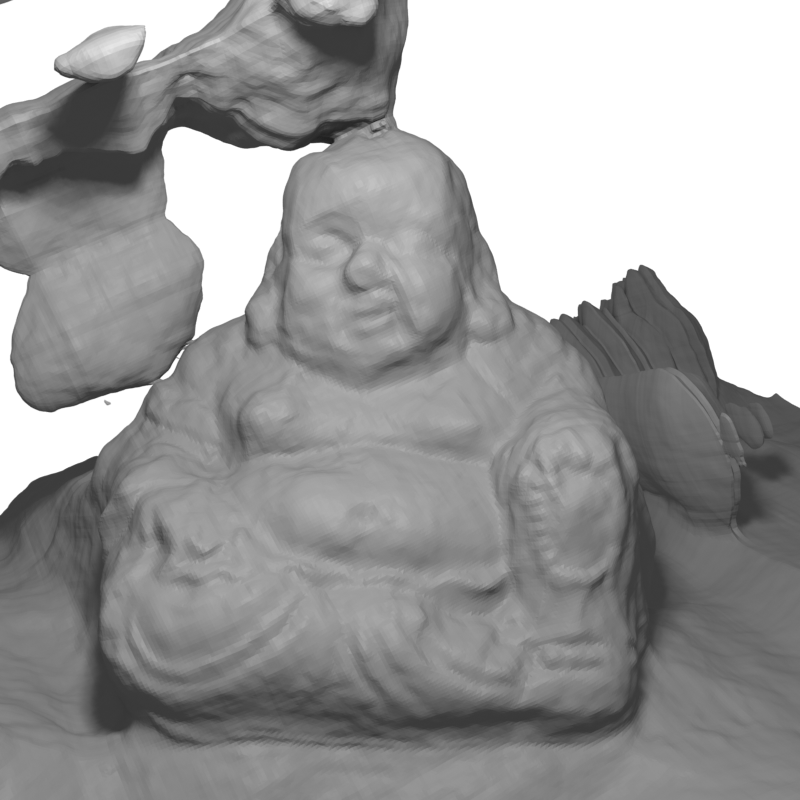} &
    \includegraphics[width=0.19\textwidth]{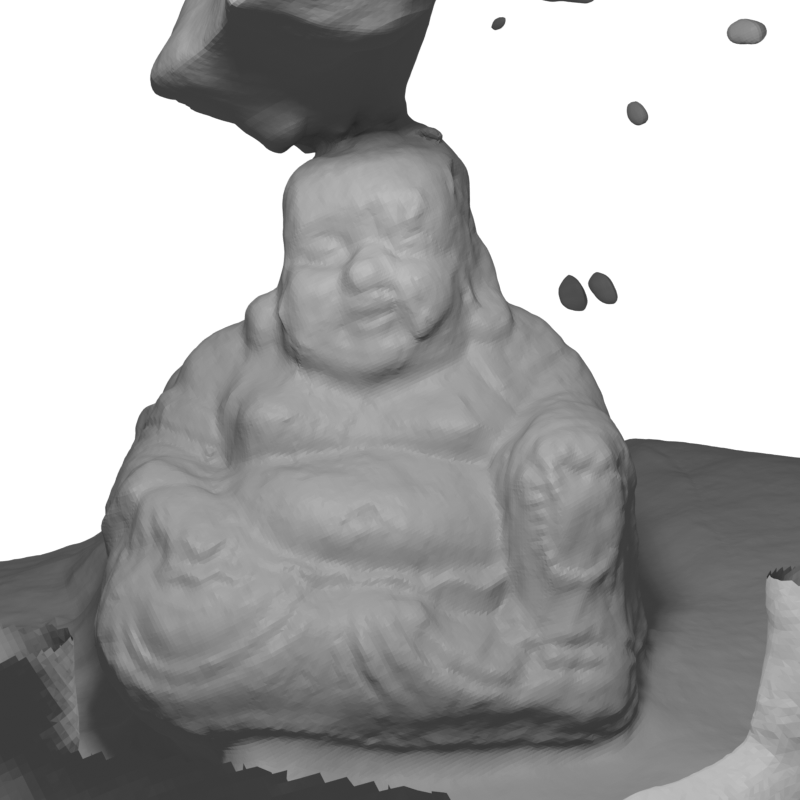}\\
    Ground Truth & MLP & Multi-Res. Grid & QFF-Lite & QFF-3D 
\end{tabular}
}
\captionof{figure}{
\textbf{Qualitative visualization of DTU dataset~\cite{aanaes2016large} reconstruction based on 3 sparse views.}
MLP and Multi-Res Grid refer to the MonoSDF~\cite{yu2022monosdf} with corresponding architectures. 
We use author provided meshes for visualizing results with MLP and Multi-Res Grid. 
} 
\label{fig:dtu_3views}
\end{table*}

\begin{table}
\centering
\resizebox{0.49\textwidth}{!}{
\begin{tabular}{l|cc}
\toprule
Method &   Chamfer-$L_1$ & \# Params \\
\midrule
TSDF-Fusion~\cite{schoenberger2016mvs}          & 4.80            &   -     \\
COLMAP~\cite{schoenberger2016mvs}               & 2.56            &   -     \\
RealityCapture                                  & 2.84            &   -     \\
MonoSDF~(Multi-Res. Grids)~\cite{yu2022monosdf} & 3.68            &  12.5 M \\
MonoSDF~(MLP)~\cite{yu2022monosdf}              & \textbf{1.86}   &  \textbf{670 K} \\
\midrule
MonoSDF~(QFF-Lite)                              & 2.15            &  837 K \\ 
MonoSDF~(QFF-3D)                                & 2.79            &  5.10 M \\ 
\bottomrule
\end{tabular}
}

\caption{
    \textbf{Evaluation on DTU dataset~\cite{aanaes2016large} with 3 views}. 
    We report the Chamfer-$L_1$ distance of baseline and our results. 
    We mark the best performing method in bold.
}
\label{tab:dtu}
\end{table}

\subsection{Neural Radiance Fields}
\label{sec:nerf}
\textbf{Dataset.} We use the NeRF Synthetic dataset~\cite{mildenhall2021nerf} to compare with existing methods. 

\textbf{Evaluation Metrics.} We evaluate the novel view synthesis performance by measuring the peak signal-to-noise ratio (PSNR).

\textbf{Implementation.} 
We apply our QFF-Lite and QFF-3D to the Neural Radiance Field  baseline NeRFAcc~\cite{li2022nerfacc}.
Table~\ref{tab:nerf} summarizes our results. 
We divide our results into decomposed scene representations, which do not explicitly compose input encodings (\eg QFF-Lite), and composed scene representations, which explicitly create a composed 3D representation (\eg QFF-3D).
For both of our methods, we use $N=16$, $L=6$ and $M=2^7$, and use Adam~\cite{kingma2014adam} optimizer with $lr=1e{-2}$ for the QFF and $lr=5e^{-4}$ for MLP.
\footnote{For scene \textit{Chair} and \textit{Ficus}, we use slightly lower learning rate of $9.5e^{-3}$ and $3.5e^{-4}$ for the QFF and the MLP due to the loss converging to NaN in NeRFAcc~\cite{li2022nerfacc}.}
We use a 4-layer MLP for QFF-Lite and a 2-layer MLP for QFF-3D for all scenes. 

\begin{table*}[t]
\centering
\begin{tabular}{l|ccccccc}
\toprule
Method &   Acc. $\downarrow$ & Comp. $\downarrow$ & Chamfer-$L_1$ $\downarrow$ & Prec. $\uparrow$ & Recall $\uparrow$ & F-score $\uparrow$ & \# Params\\
\midrule
COLMAP~\cite{schoenberger2016mvs}               & 0.047          & 0.235          & 0.141          & 0.711          & 0.441          & 0.537            & -         \\
UNISURF~\cite{oechsle2021unisurf}               & 0.554          & 0.164          & 0.359          & 0.212          & 0.362          & 0.267            & 802 K     \\ 
NeuS~\cite{wang2021neus}                        & 0.179          & 0.208          & 0.194          & 0.313          & 0.275          & 0.291            & 1.41 M    \\ 
VolSDF~\cite{yariv2021volume}                   & 0.414          & 0.120          & 0.267          & 0.321          & 0.394          & 0.346            & 802 K     \\ 
Manhattan-SDF~\cite{guo2022neural}              & 0.072          & 0.068          & 0.070          & 0.621          & 0.586          & 0.602            & 1.06 M    \\ 
NeuRIS~\cite{wang2022neuris}                    & 0.050          & 0.049          & 0.050          & 0.717          & 0.669          & 0.692            & 1.41 M    \\ 
MonoSDF~(Multi-Res. Grids)~\cite{yu2022monosdf} & 0.072          & 0.057          & 0.064          & 0.770          & 0.601          & 0.626            & 12.5 M    \\ 
MonoSDF~(MLP)~\cite{yu2022monosdf}              & \textbf{0.035} & 0.048          & 0.042          & \textbf{0.799} & 0.681          & 0.733            & \textbf{711 K}     \\ 
\midrule
MonoSDF~(QFF-Lite)                              & 0.043          & 0.044          & 0.044          & 0.761          & 0.718          & 0.738            & 1.63 M    \\ 
MonoSDF~(QFF-3D)                                & 0.040          & \textbf{0.041} & \textbf{0.041} & 0.765          & \textbf{0.744} & \textbf{0.754}   & 9.97 M    \\ 
\bottomrule
\end{tabular}

\caption{
    \textbf{Evaluation on ScanNet~\cite{dai2017scannet}}. We report the baseline and our results on ScanNet~\cite{dai2017scannet}. We mark bold for the best methods for each criteria. We use the results provided by~\cite{yu2022monosdf} for the baseline.
}
\label{tab:scannet}
\end{table*}

\begin{table*}[t!]
\resizebox{\textwidth}{!}{
\centering
\begin{tabular}{ccccc}
    {\tikz{
    \node[draw=black, draw opacity=1.0, line width=.3mm, inner sep=0pt](qff) at (0, 0) {\includegraphics[width=0.19\textwidth]{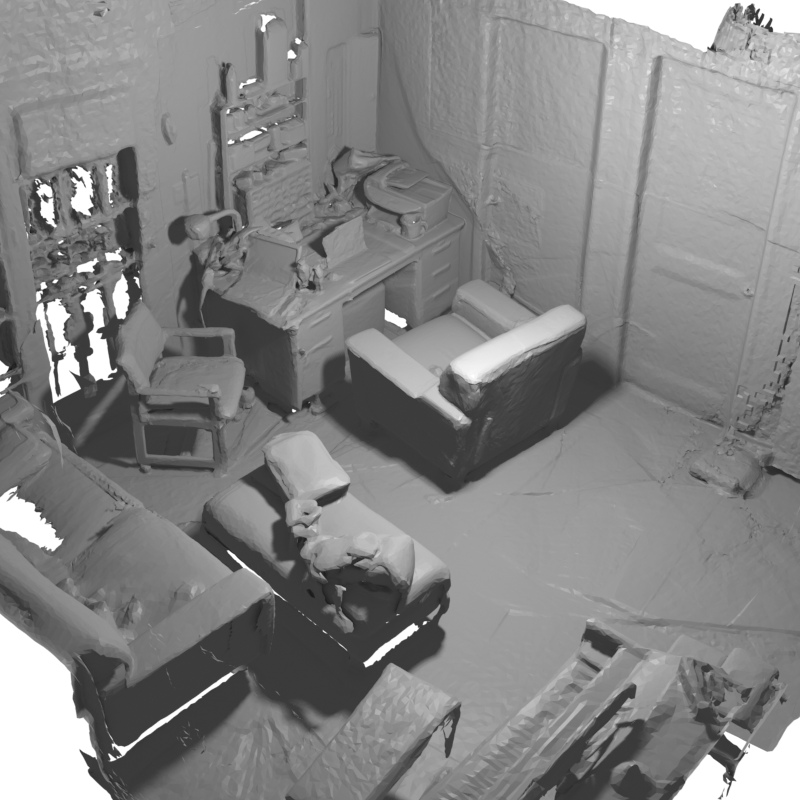}};
    \node[draw=black, draw opacity=1.0, line width=.3mm, inner sep=0pt](qff) at (-.9, -.9) {\includegraphics[trim=6cm 14cm 18cm 10cm, clip, width=0.08\textwidth]{images/scannet/scan_1_gt_side.png}};
    }}
     &
    {\tikz{
    \node[draw=black, draw opacity=1.0, line width=.3mm, inner sep=0pt](qff) at (0, 0) {\includegraphics[width=0.19\textwidth]{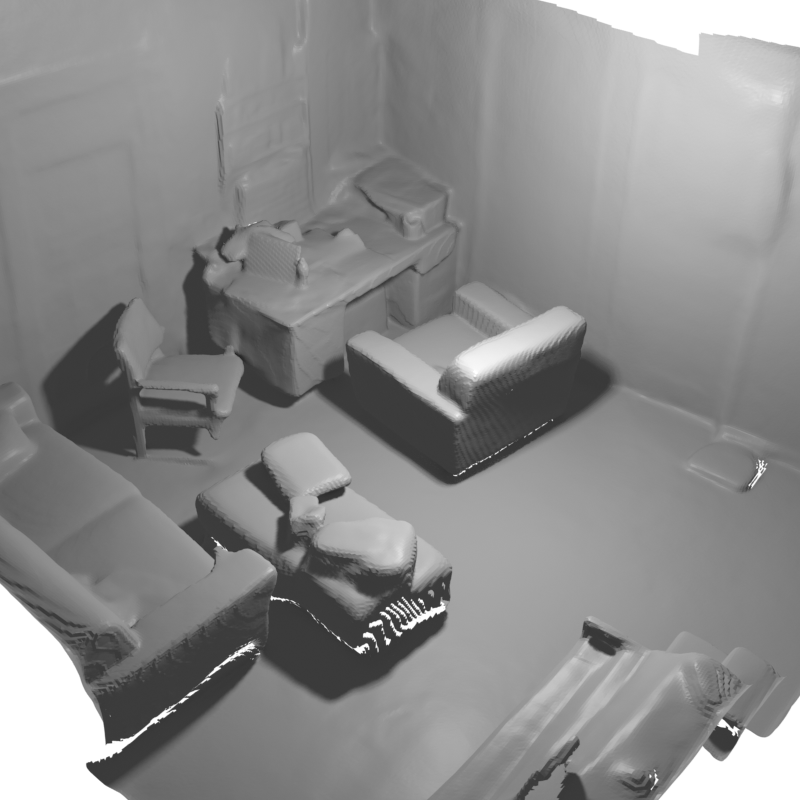}};
    \node[draw=black, draw opacity=1.0, line width=.3mm, inner sep=0pt](qff) at (-.9, -.9) {\includegraphics[trim=6cm 14cm 18cm 10cm, clip, width=0.08\textwidth]{images/scannet/scan_1_mlp_side.png}};
    }} & 
    {\tikz{
    \node[draw=black, draw opacity=1.0, line width=.3mm, inner sep=0pt](qff) at (0, 0) {\includegraphics[width=0.19\textwidth]{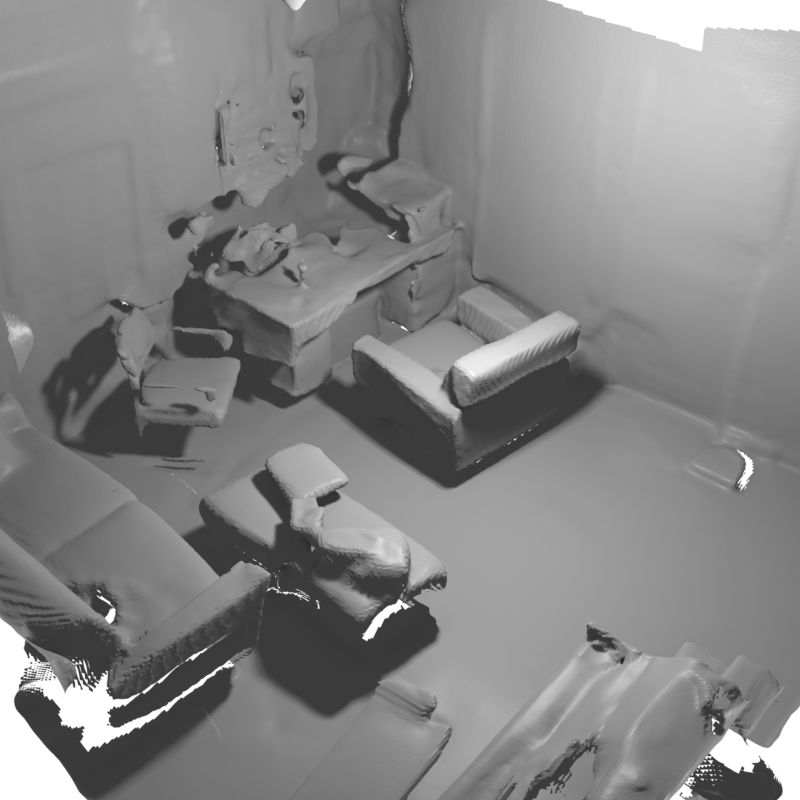}};
    \node[draw=black, draw opacity=1.0, line width=.3mm, inner sep=0pt](qff) at (-.9, -.9) {\includegraphics[trim=6cm 14cm 18cm 10cm, clip, width=0.08\textwidth]{images/scannet/scan_1_grid_side.png}};
    }} & 
    {\tikz{
    \node[draw=black, draw opacity=1.0, line width=.3mm, inner sep=0pt](qff) at (0, 0) {\includegraphics[width=0.19\textwidth]{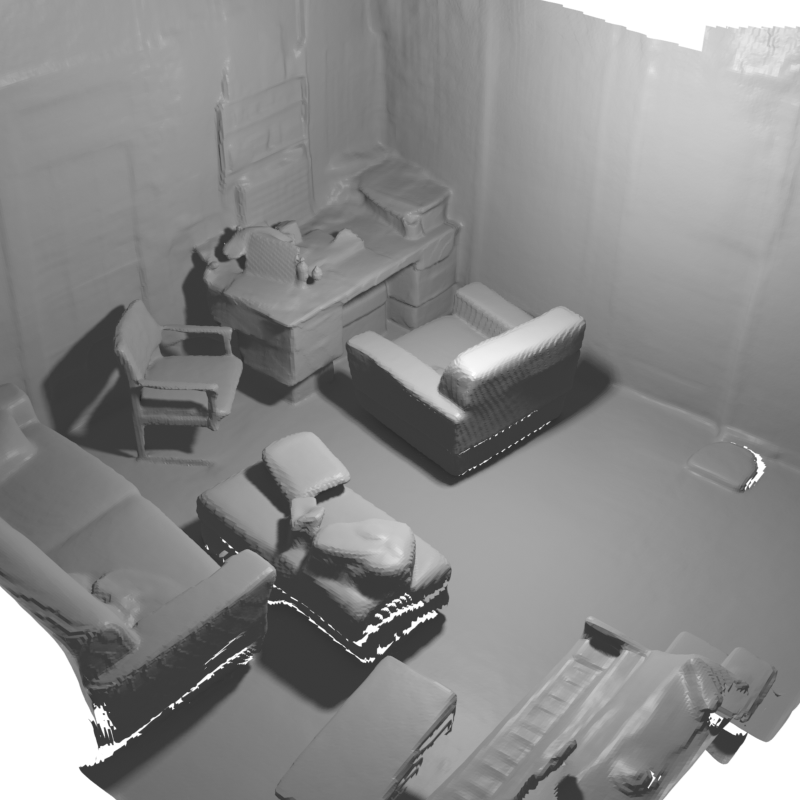}};
    \node[draw=black, draw opacity=1.0, line width=.3mm, inner sep=0pt](qff) at (-.9, -.9) {\includegraphics[trim=6cm 14cm 18cm 10cm, clip, width=0.08\textwidth]{images/scannet/scan_1_qff_lite_side.png}};
    }} & 
    {\tikz{
    \node[draw=black, draw opacity=1.0, line width=.3mm, inner sep=0pt](qff) at (0, 0) {\includegraphics[width=0.19\textwidth]{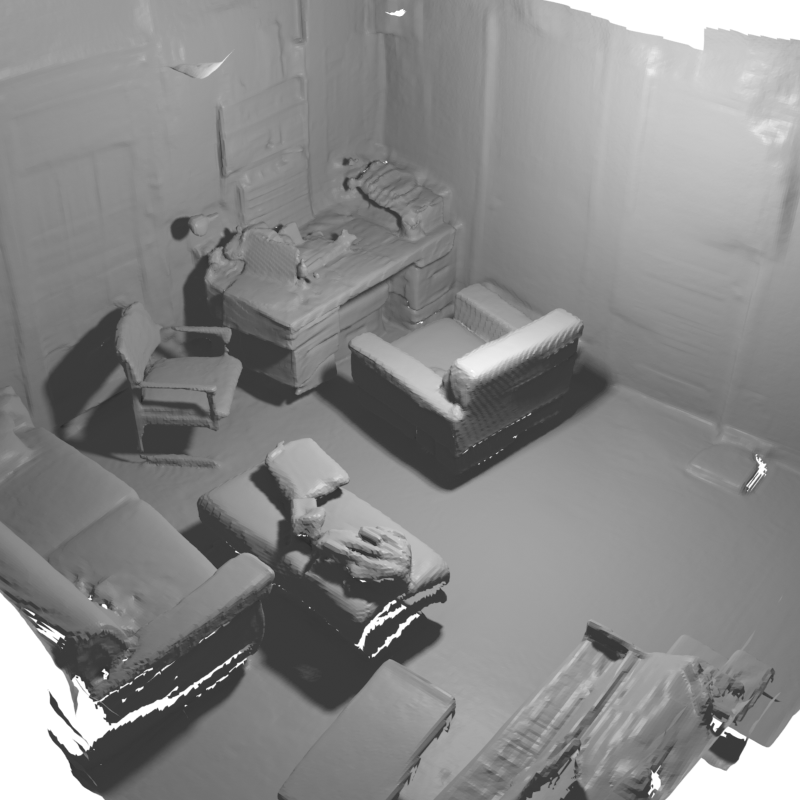}};
    \node[draw=black, draw opacity=1.0, line width=.3mm, inner sep=0pt](qff) at (-.9, -.9) {\includegraphics[trim=6cm 14cm 18cm 10cm, clip, width=0.08\textwidth]{images/scannet/scan_1_qff_3d_side.png}};
    }} 
    \\
    {\tikz{
    \node[draw=black, draw opacity=1.0, line width=.3mm, inner sep=0pt](qff) at (0, 0) {\includegraphics[width=0.19\textwidth]{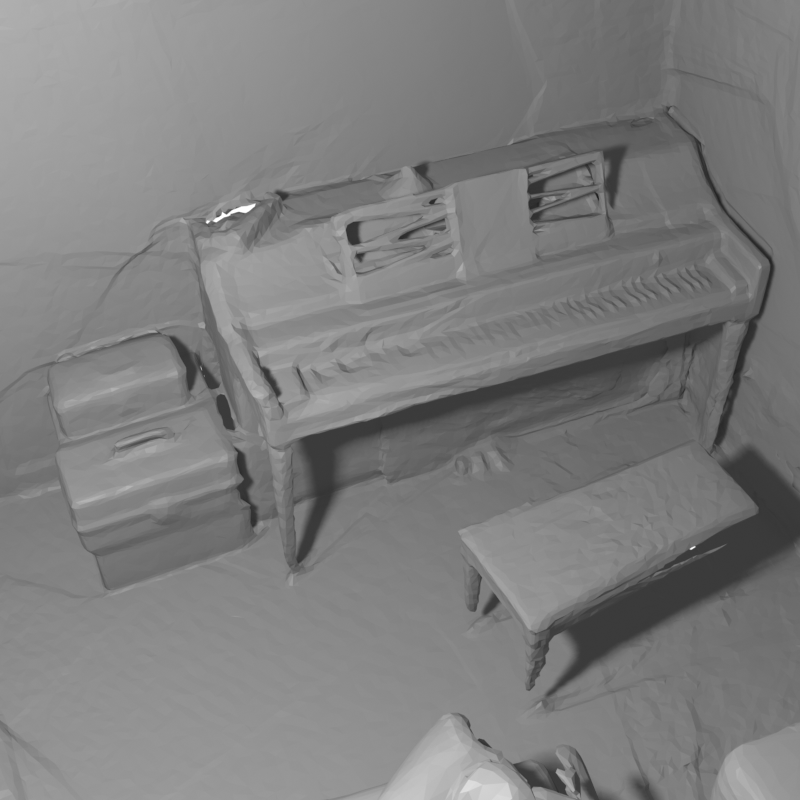}};
    \node[draw=black, draw opacity=1.0, line width=.3mm, inner sep=0pt](qff) at (-.9, -.9) {\includegraphics[trim=15.5cm 5.5cm 5.5cm 15.5cm, clip, width=0.08\textwidth]{images/scannet/scan_1_gt_piano.png}};
    }} & 
    {\tikz{
    \node[draw=black, draw opacity=1.0, line width=.3mm, inner sep=0pt](qff) at (0, 0) {\includegraphics[width=0.19\textwidth]{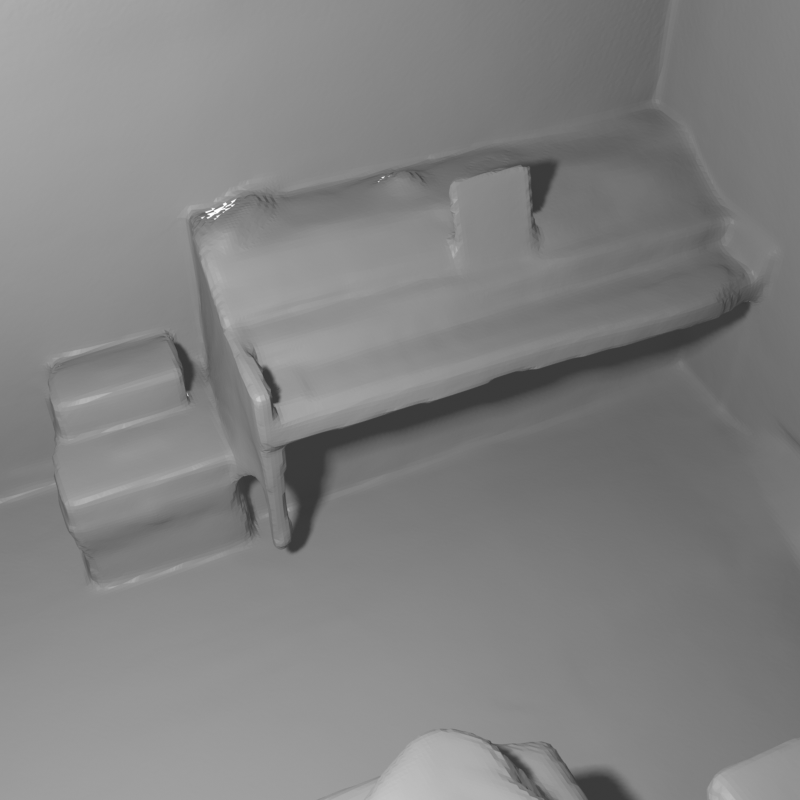}};
    \node[draw=black, draw opacity=1.0, line width=.3mm, inner sep=0pt](qff) at (-.9, -.9) {\includegraphics[trim=15.5cm 5.5cm 5.5cm 15.5cm, clip, width=0.08\textwidth]{images/scannet/scan_1_mlp_piano.png}};
    }} & 
    {\tikz{
    \node[draw=black, draw opacity=1.0, line width=.3mm, inner sep=0pt](qff) at (0, 0) {\includegraphics[width=0.19\textwidth]{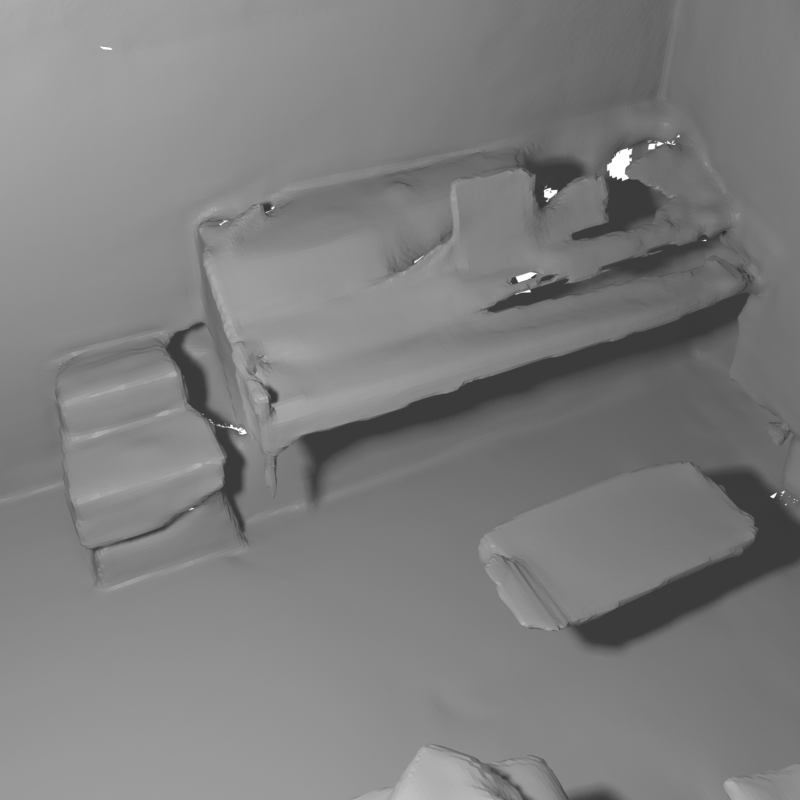}};
    \node[draw=black, draw opacity=1.0, line width=.3mm, inner sep=0pt](qff) at (-.9, -.9) {\includegraphics[trim=15.5cm 5.5cm 5.5cm 15.5cm, clip, width=0.08\textwidth]{images/scannet/scan_1_grid_piano.png}};
    }} & 
    {\tikz{
    \node[draw=black, draw opacity=1.0, line width=.3mm, inner sep=0pt](qff) at (0, 0) {\includegraphics[width=0.19\textwidth]{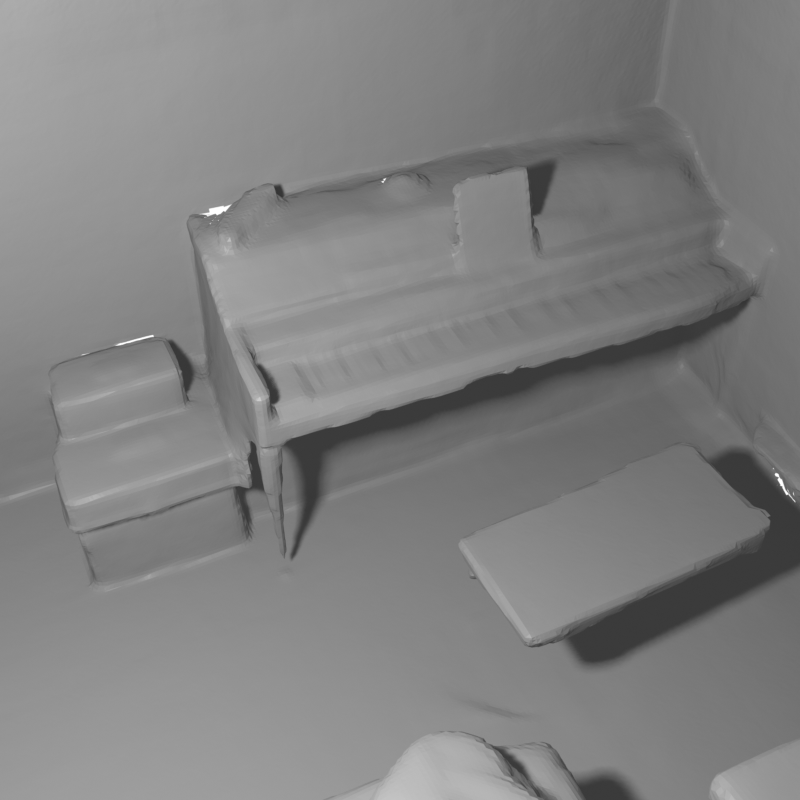}};
    \node[draw=black, draw opacity=1.0, line width=.3mm, inner sep=0pt](qff) at (-.9, -.9) {\includegraphics[trim=15.5cm 5.5cm 5.5cm 15.5cm, clip, width=0.08\textwidth]{images/scannet/scan_1_qff_lite_piano.png}};
    }} & 
    {\tikz{
    \node[draw=black, draw opacity=1.0, line width=.3mm, inner sep=0pt](qff) at (0, 0) {\includegraphics[width=0.19\textwidth]{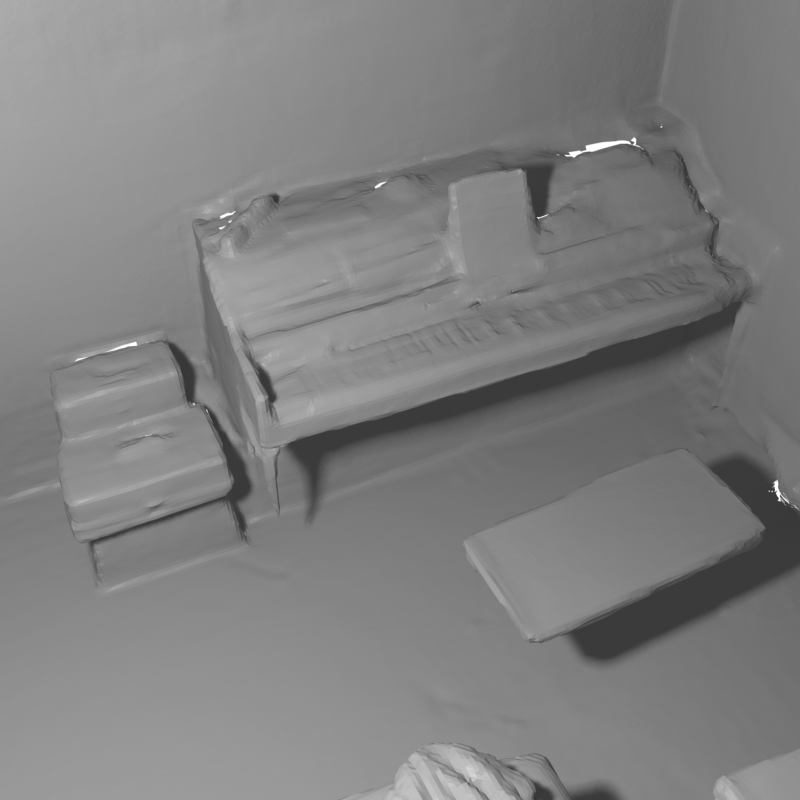}};
    \node[draw=black, draw opacity=1.0, line width=.3mm, inner sep=0pt](qff) at (-.9, -.9) {\includegraphics[trim=15.5cm 5.5cm 5.5cm 15.5cm, clip, width=0.08\textwidth]{images/scannet/scan_1_qff_3d_piano.png}};
    }} \\
    {\tikz{
    \node[draw=black, draw opacity=1.0, line width=.3mm, inner sep=0pt](qff) at (0, 0) {\includegraphics[width=0.19\textwidth]{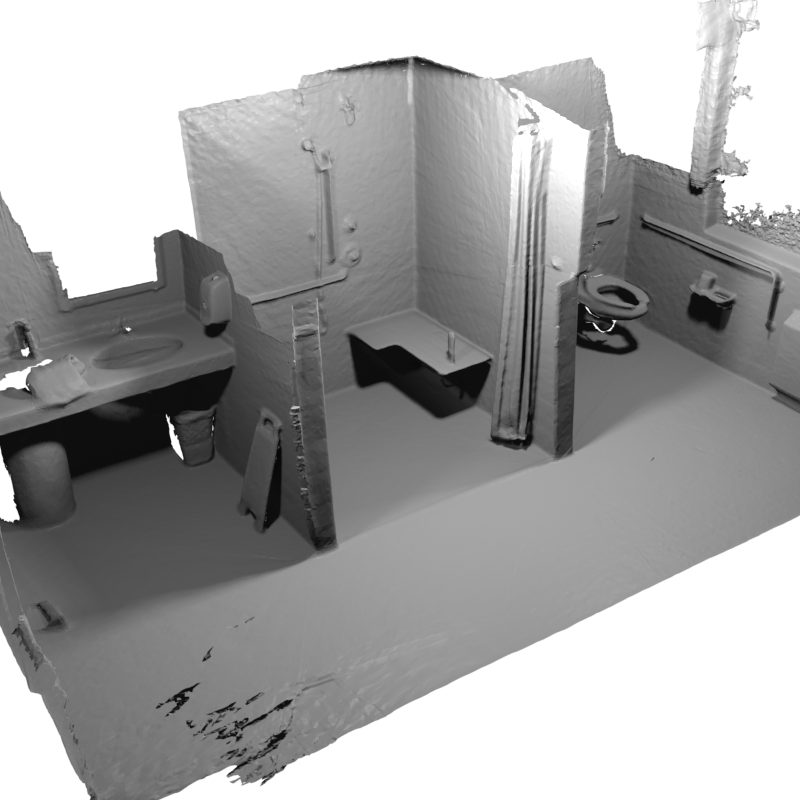}};
    \node[draw=black, draw opacity=1.0, line width=.3mm, inner sep=0pt](qff) at (-.9, -.9) {\includegraphics[trim=5cm 13cm 17cm 9cm, clip, width=0.08\textwidth]{images/scannet/scan_2_gt_side.png}};
    }} & 
    {\tikz{
    \node[draw=black, draw opacity=1.0, line width=.3mm, inner sep=0pt](qff) at (0, 0) {\includegraphics[width=0.19\textwidth]{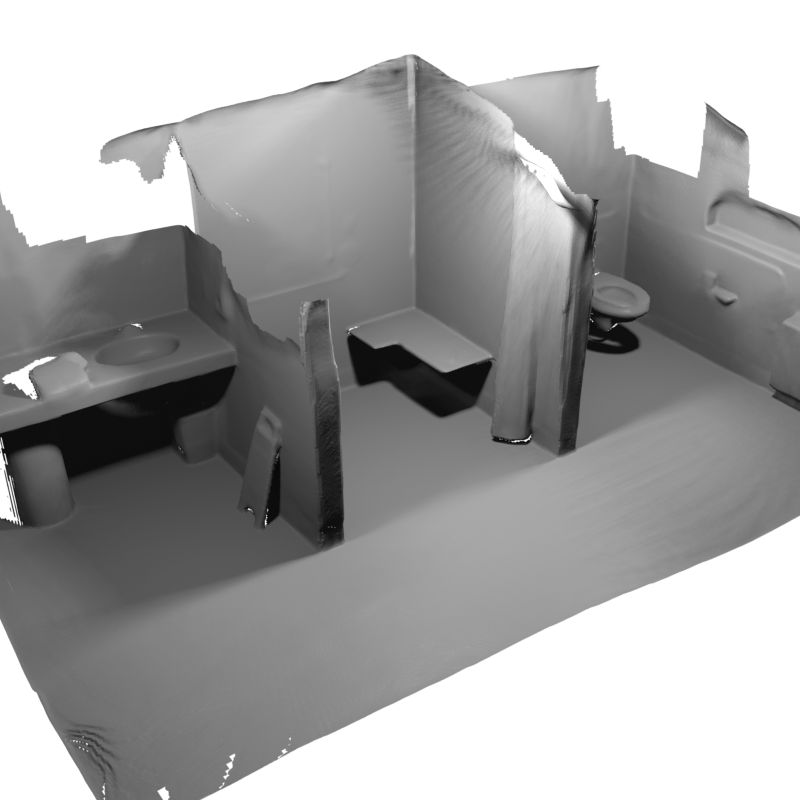}};
    \node[draw=black, draw opacity=1.0, line width=.3mm, inner sep=0pt](qff) at (-.9, -.9) {\includegraphics[trim=5cm 13cm 17cm 9cm, clip, width=0.08\textwidth]{images/scannet/scan_2_mlp_side.png}};
    }} & 
    {\tikz{
    \node[draw=black, draw opacity=1.0, line width=.3mm, inner sep=0pt](qff) at (0, 0) {\includegraphics[width=0.19\textwidth]{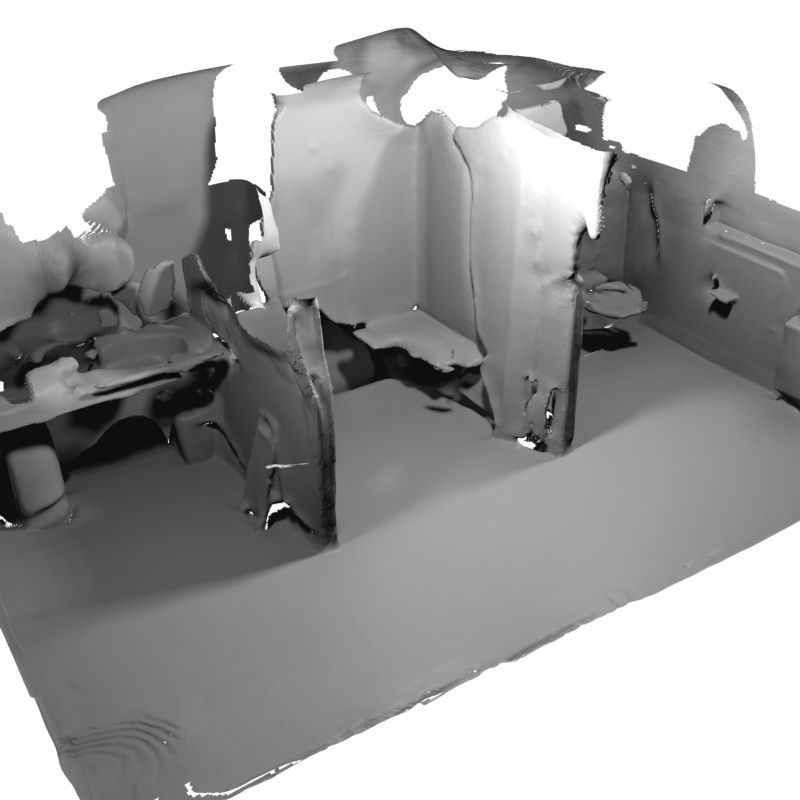}};
    \node[draw=black, draw opacity=1.0, line width=.3mm, inner sep=0pt](qff) at (-.9, -.9) {\includegraphics[trim=5cm 13cm 17cm 9cm, clip, width=0.08\textwidth]{images/scannet/scan_2_grid_side.png}};
    }} & 
    {\tikz{
    \node[draw=black, draw opacity=1.0, line width=.3mm, inner sep=0pt](qff) at (0, 0) {\includegraphics[width=0.19\textwidth]{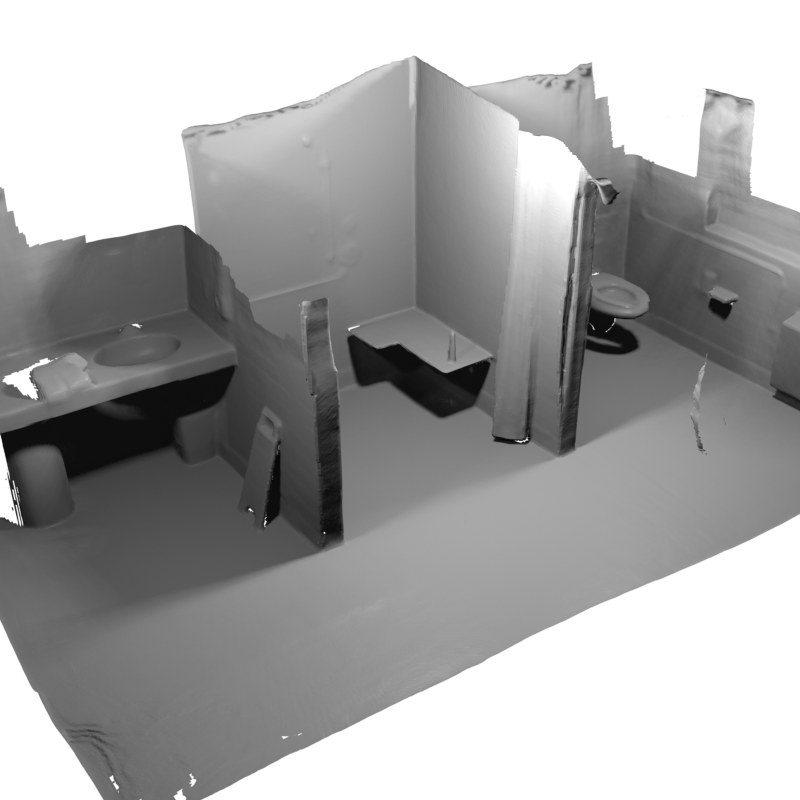}};
    \node[draw=black, draw opacity=1.0, line width=.3mm, inner sep=0pt](qff) at (-.9, -.9) {\includegraphics[trim=5cm 13cm 17cm 9cm, clip, width=0.08\textwidth]{images/scannet/scan_2_qff_lite_side.png}};
    }} & 
    {\tikz{
    \node[draw=black, draw opacity=1.0, line width=.3mm, inner sep=0pt](qff) at (0, 0) {\includegraphics[width=0.19\textwidth]{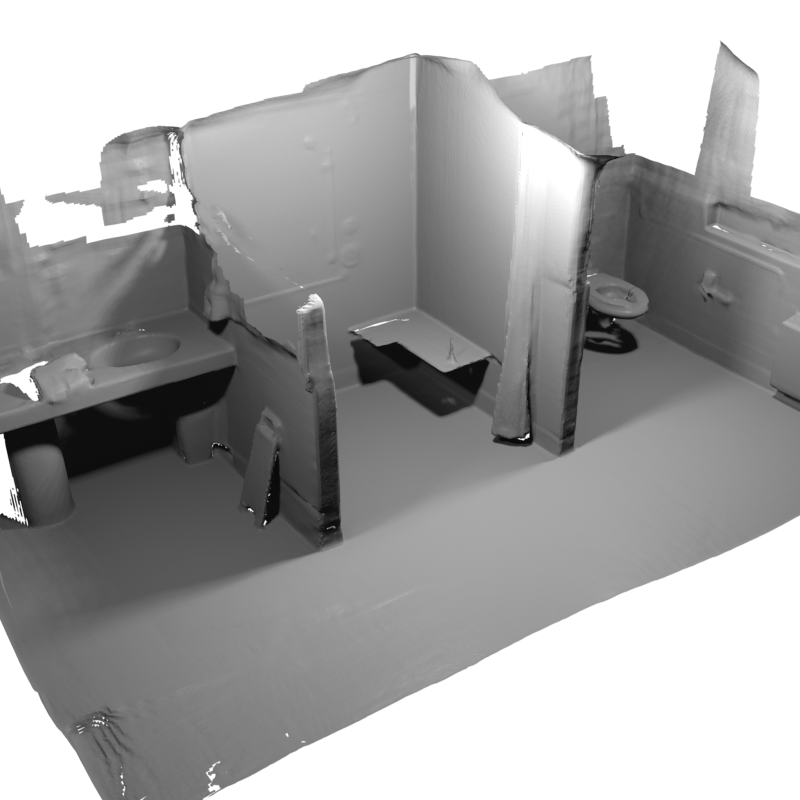}};
    \node[draw=black, draw opacity=1.0, line width=.3mm, inner sep=0pt](qff) at (-.9, -.9) {\includegraphics[trim=5cm 13cm 17cm 9cm, clip, width=0.08\textwidth]{images/scannet/scan_2_qff_3d_side.png}};
    }} \\
    Ground Truth & MLP & Multi-Res. Grid & QFF-Lite & QFF-3D 
\end{tabular}
}

\captionof{figure}{
    \textbf{Qualitative visualization of ScanNet Reconstruction.}
    MLP and Multi-Res Grid refers to the MonoSDF~\cite{yu2022monosdf} with corresponding architectures. 
    We use author provided meshes for visualizing the MLP, Multi-Res Grid. Best viewed in zoomed. 
}

\label{fig:scannet}
\end{table*}
\vspace{1em}
\textbf{Results.} As shown in Table~\ref{tab:nerf}, in methods using decomposed scene representations, 
our QFF-Lite achieves the highest overall PSNR with smaller number of parameters, compared to all baseline methods.
For methods with composed scene representations, our QFF-3D results are mostly on par with existing methods, with PSNR differences of at most 0.1 for 3 out of 8 scenes, and achieve slightly higher average PSNR with a smaller number of parameters.
We emphasize that we do not impose a total variation~(TV) loss or have to crop and resize the feature vectors as in TensoRF~\cite{chen2022tensorf}. This is because our representation adds original positional encoding to preserve smoothness and is naturally multiscale and shift-invariant.

We present our qualitative comparisons in Figure~\ref{fig:nerf_synthetic}. In both composed and decomposed scene representations, our model is able to capture high frequency details, such as inner-microphone geometry and Lego textures, with larger PSNR improvements compared to baseline models.

\subsection{Neural Signed Distance Fields}
\label{sec:nsdf}
\textbf{Datasets.}
We use DTU dataset~\cite{aanaes2016large} for sparse view (3-view) object reconstruction and ScanNet dataset~\cite{dai2017scannet} for large-scale reconstruction.

\textbf{Evaluation Metrics.}
For the DTU dataset ~\cite{aanaes2016large}, we measure the Chamfer distance using the evaluation protocol provided by the dataset.
For the ScanNet dataset, we report the Chamfer distance and $F_1$ score following the baseline methods~\cite{guo2022neural,yu2022monosdf}.

\textbf{Sparse view object reconstruction}
We apply our QFF-Lite and QFF-3D on MonoSDF~\cite{yu2022monosdf} for DTU dataset~\cite{aanaes2016large} with three views. 
We use $N=8$, $M=2^7$ $L=6$,  8-layer MLP for QFF-Lite and 2-layer MLP for QFF-3D for all scenes.
Table~\ref{tab:dtu} shows the results of our methods compared to the baselines. We find that both our QFF-Lite and QFF-3D are better than MonoSDF with Multi-Resolution Grids, but worse than MonoSDF with MLP. Some artifacts appear in portions of the scene that are not well observed by the three views.  We find our results consistent with those in MonoSDF: MLP performs better due to larger smoothness bias in sparse views, and the high-parameter models tend to overfit or produce spurious artifacts. Qualitatively, shown in Figure~\ref{fig:dtu_3views}, our method can reconstruct thin regions, such as the arms and noses of the small snowman, which MLP cannot complete. MLP captures better geometry in smooth regions with high frequency textures, such as faces.

\textbf{Large scale scene reconstruction.}
Similarly, we apply our QFF-Lite and QFF-3D on MonoSDF~\cite{yu2022monosdf} for the ScanNet dataset~\cite{dai2017scannet}.
We use $N=16$, $M=2^7$ $L=6$,  8-layer MLP for QFF-Lite and 2-layer MLP for QFF-3D for all scenes.
Table~\ref{tab:scannet} compares ScanNet evaluation to the baseline methods. 
Our QFF-3D achieves the best Chamfer-$L_1$ distance, and the best overall F-score, followed by our QFF-Lite.
In more detail, we achieve significantly higher recall but slightly worse accuracy and precision compared to MLP and Multi-Res Grid.
We also qualitatively verify the robustness of our method as shown in Figure~\ref{fig:scannet}. 
Our QFF-Lite and QFF-3D are capable of capturing correct geometry compared to Multi-Res Grid, and are on-par with MLP. We also emphasize that both of our methods are capable of capturing high-frequency geometries, such as piano keyboards or drawer hands.

\subsection{Ablation Studies and Discussions}
We compare the impact of different design choices of our method on the \textit{Lego} scenario of the Nerf Synthetic Dataset~\cite{mildenhall2021nerf}. 
We use the default values of both QFF-Lite and QFF-3D with $N=16$, $M=2^7$ and $L=6$, and use 8-layer MLP for QFF-Lite and 2-layer MLP for QFF-3D.
Table~\ref{tab:ablation} shows a summary of our ablation studies. 

\noindent\textbf{Resolution of Feature Bins.}
Given the same feature vector length and number of layers, for QFF-Lite, we find a small increment of PSNR as the number of bins increases, but the increment slows down as we further increase the resolution, with 0.24 PSNR improvement from $M=2^5 \rightarrow 2^7$ and 0.04 PSNR from $M=2^7 \rightarrow 2^9$.
For QFF-3D, increasing the number of bins to an excessively high number reduces the PSNR, because each bin in 2D is too fine-grained to be accessed multiple times. 
The number of parameters is proportional to $\mathcal{O}(M)$ for QFF-Lite and $\mathcal{O}(M^2)$ for QFF-3D.

\begin{table}[t]
\centering
\resizebox{0.49\textwidth}{!}{
\begin{tabular}{@{}l| ccc |cc@{}}
\toprule
Method          & Feats.~(N) & Bins~(M) & Layers & PSNR & \# Params \\
\midrule
\multirow{5}{*}{QFF-Lite}
& \textit{8}    & $2^7$          & 8            & 35.41 & 786 K \\
& \textbf{16}   & $\mathbf{2^7}$ & \textbf{8}   & 35.52 & 934 K \\
& \textit{32}   & $2^7$          & 8            & 35.83 & 1.45 M \\
& 16            & $\mathit{2^5}$ & 8            & 35.28 & 897 K \\
& 16            & $\mathit{2^9}$ & 8            & 35.56 & 1.45 M \\
& 16            & $2^7$          & \textit{6}   & 35.56 & 876 K \\
& 16            & $2^7$          & \textit{10}  & 35.02 & 1.28 M \\
\midrule
\multirow{7}{*}{QFF-3D} 
& \textit{8}    & $2^7$          & 2            & 36.22 & 5.00 M \\
& \textbf{16}   & $\mathbf{2^7}$ & \textbf{2}   & 36.68 & 9.82 M \\
& \textit{32}   & $2^7$          & 2            & 36.91 & 19.5 M \\
& 16            & $\mathit{2^5}$ & 2            & 35.28 & 925 K \\
& 16            & $\mathit{2^9}$ & 2            & 36.00 & 151 M \\
& 16            & $2^7$          & \textit{3}   & 36.83 & 9.89 M \\
& 16            & $2^7$          & \textit{4}   & 36.82 & 9.96 M \\
\bottomrule
\end{tabular}
}
\caption{
    \textbf{Comparison of different hyper-parameters.} \textbf{Bold} values denote our default model. \textit{Italicized} values denote the changes to the default model. 
}
\label{tab:ablation}
\end{table}
\vspace{1em}
\begin{figure}[t]
    \centering
    \includegraphics[width=0.49\textwidth]{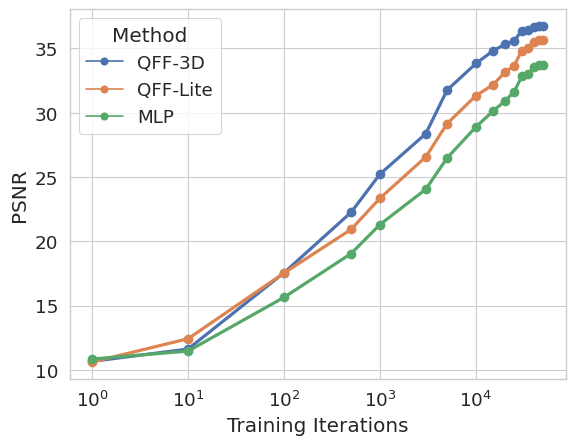}
    \caption{
        \textbf{Comparison of Convergence for the \textit{Lego} scene.} 
        We visualize number of steps taken vs. Test Image PSNR for our \textit{QFF-Lite} and \textit{QFF-3D} against MLP.
    }
    \label{fig:ablation_convergence}
\end{figure}
\vspace{1em}
\noindent\textbf{Length of Feature Vector.}
We find that increasing the length of the feature vectors benefits both QFF-Lite and QFF-3D, improving 0.11, 0.46 PSNRs from $N=8 \rightarrow 16$, and 0.31, 0.23 PSNRs from $N=16 \rightarrow 32$, for QFF-Lite and QFF-3D respectively. 
Changing $N$ changes the number of parameters proportional to $\mathcal{O}(M)$ for QFF-Lite and $\mathcal{O}(M^2)$ for QFF-3D.

\noindent\textbf{Number of Layers.}
We vary the number of layers in QFF-3D and find a slight improvement of 0.15 PSNR with MLP going from 2 to 3 layers, but 4-layer does not improve further.

\begin{figure}[t]
    \centering
    \begin{tabular}{@{}c@{}c@{}}
        \includegraphics[width=0.24\textwidth]{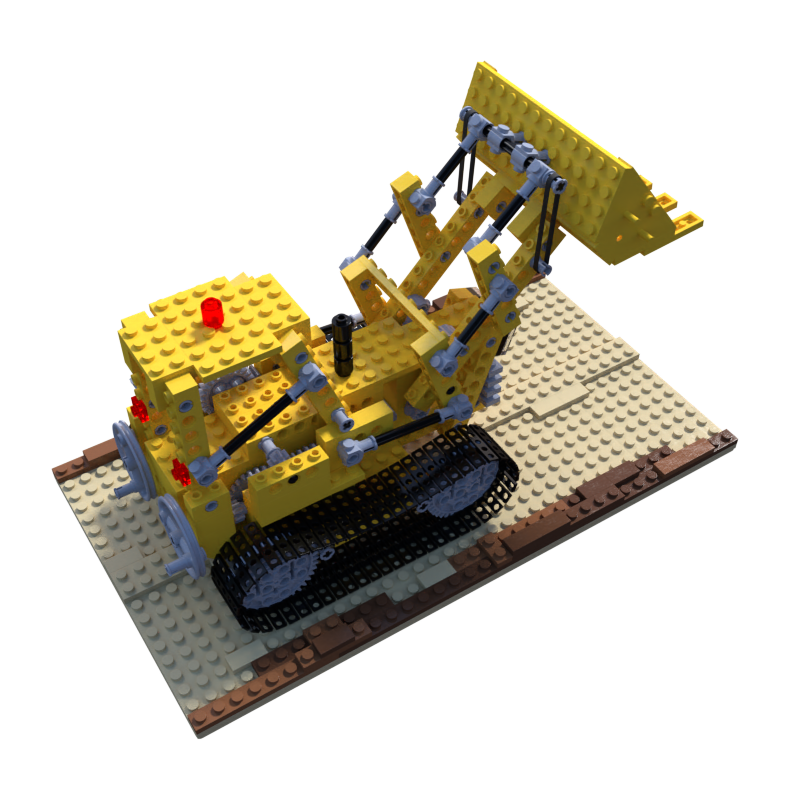} & 
        \includegraphics[width=0.24\textwidth]{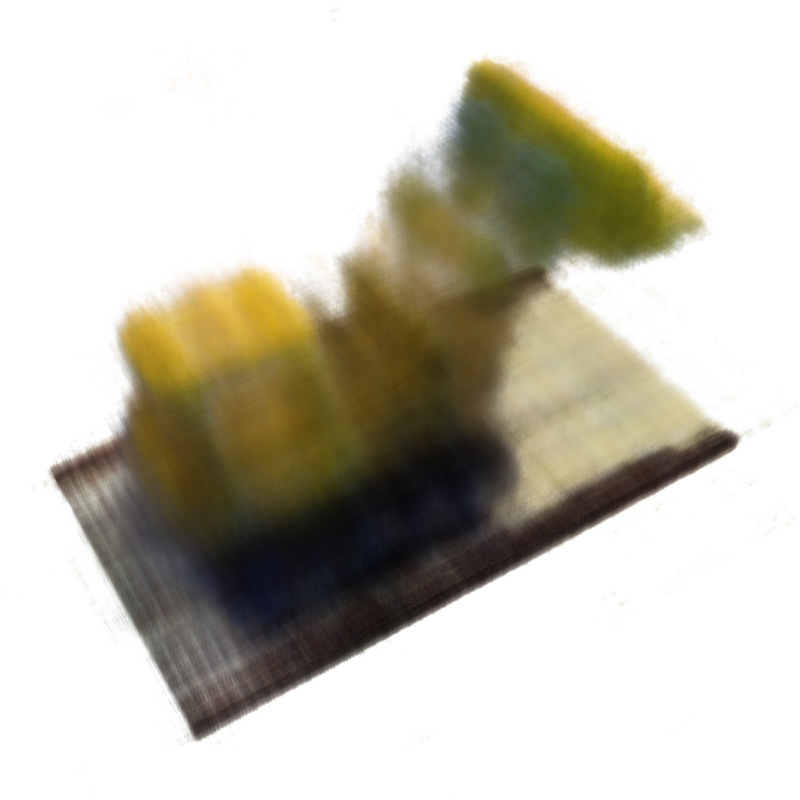} \\
        Ground truth & NeRFAcc\cite{li2022nerfacc} \\
        \includegraphics[width=0.24\textwidth]{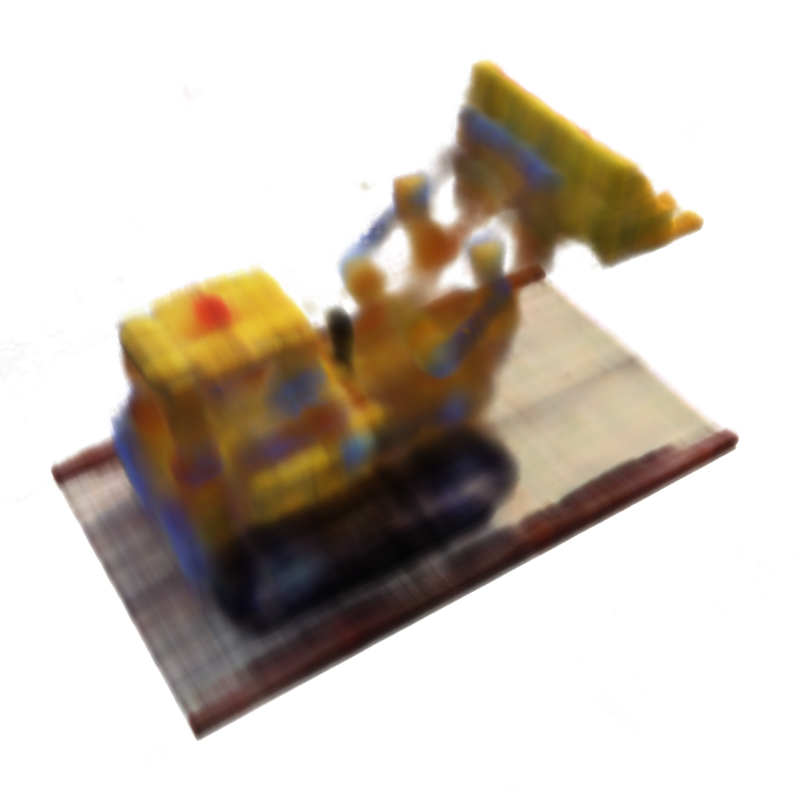} & 
        \includegraphics[width=0.24\textwidth]{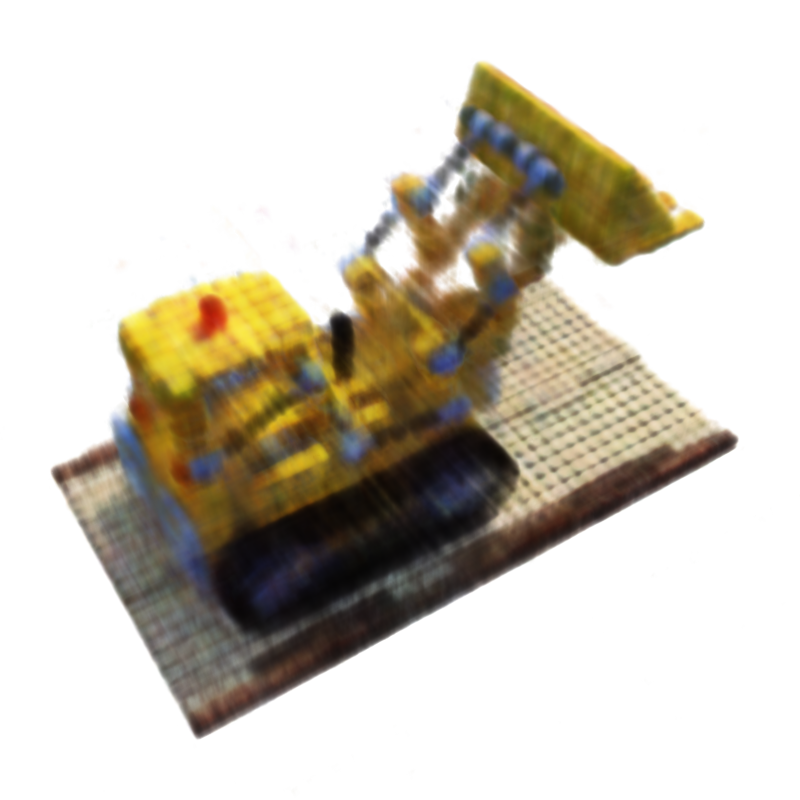} \\
        NeRFAcc (QFF-Lite) & NeRFAcc (QFF-3D)
    \end{tabular}
    \caption{
        \textbf{Visualization of the \textit{Lego} scene at 500 steps of training.} Best viewed in zoomed. 
    }
    \label{fig:ablation_convergence_vis}
\end{figure}

\noindent\textbf{Convergence time compared to MLP.}
In addition, we compare the convergence speed of the test-time loss between our method and the baseline 8-layer MLP. 
Figure~\ref{fig:ablation_convergence} shows the plot of the convergence time of ours and the MLP. 
We show that our QFF-3D trains the fastest, followed by QFF-Lite, then MLP. 
We emphasize that the first 5000 iterations of our QFF-3D roughly correspond to 15000 iterations of QFF-Lite and 25000 iterations of MLP.
Figure~\ref{fig:ablation_convergence_vis} visualizes the MLP, QFF-Lite and QFF-3D only after 500 iterations of training. We show that both our QFF-Lite and QFF-3D produce sharper renderings over MLP, and that QFF-3D generates high-frequency details within a few iterations. 

\vspace{1em}
\vspace{-1.5em}
\section{Conclusion}
We present Quantized Fourier Features~(QFF), an easy-to-train, memory-efficient yet expressive neural field representation. QFF combines the best worlds of explicit feature representation and frequency-based positional encoding. We demonstrate advantages of QFF in wide range of applications of neural field representations.

\noindent\textbf{Acknowledgements.} This research is partially supported by NSF IIS 2020227, a gift from Amazon and a gift from Illinois-Insper Collaborative Research Fund.


{\small
\bibliographystyle{ieee_fullname}
\bibliography{bibliography}
}

\end{document}